\documentclass[10pt,twocolumn,letterpaper]{article}

\usepackage{iccv}      % To produce the REVIEW version
\definecolor{iccvblue}{rgb}{0.21,0.49,0.74}
\usepackage[pagebackref,breaklinks,colorlinks,allcolors=iccvblue]{hyperref}

\usepackage{pifont}
\usepackage{color}
\usepackage{xcolor}
\definecolor{tablered}{rgb}{1.00, 0.90, 0.95} % 定义tablered颜色为RGB值
\definecolor{tableyellow}{rgb}{1.0, 0.98, 0.8} % 淡黄色 
\usepackage{nicefrac}       %
\usepackage{multirow}
\usepackage{multicol}
\usepackage{dsfont}
\usepackage{tabu}                     % 表格插入
\usepackage{float}                    % 图片浮动
\usepackage{makecell}                 % 三线表-竖线
\usepackage{booktabs}                 % 三线表-短细横线
\usepackage{colortbl}
\usepackage[percent]{overpic}

\def\paperID{7103} % *** Enter the Paper ID here
\def\confName{ICCV}
\def\confYear{2025}

\title{Selective Structured State Space for Multispectral-fused Small Target Detection}

\author{Qianqian Zhang$^{1,2}$\footnotemark[1]\
\quad
Weijun Wang$^3$
\quad
Zihan Wang$^{5}$
\quad
Li Zhou$^{1}$
\quad
Hao Zhao$^{3}$
\quad
\\
Junshe An$^{1,4}$
\quad
Yunxing Liu$^3$\\
$^1$National Space Science Center, Chinese Academy of Sciences, China \\ \quad $^2$School of Computer Science and Technology, University of Chinese Academy of Sciences, China \\ \quad $^3$Institute for AI Industry Research (AIR), Tsinghua University, China \\ \quad $^4$School of Astronomy and Space Science, University of Chinese Academy of Sciences, China\\
\quad $^5$School of Computing, National University of Singapore\\
{\tt\small zhangqianqian21@mails.ucas.ac.cn} \quad {\tt\small wangweijun@air.tsinghua.edu.cn} \quad {\tt\small zhouli@nssc.ac.cn}  \\{\tt\small liuyunxin@air.tsinghua.edu.cn} \quad {\tt\small zhaohao@air.tsinghua.edu.cn} \quad {\tt\small anjunshe@nssc.ac.cn}\\ {\tt\small zihan.wang@u.nus.edu} 
}
\begin{document}

\maketitle
\renewcommand{\thefootnote}{\fnsymbol{footnote}} %将脚注符号设置为fnsymbol类型，即特殊符号表示
% \footnotetext{This paper has been accepted by NSDI.}
\footnotetext[1]{This work was completed during Qianqian Zhang's research internship at the Institute for AI Industry Research (AIR), Tsinghua University.} %对应脚注[1]

\begin{abstract}
Target detection in high-resolution remote sensing imagery faces challenges due to the low recognition accuracy of small targets and high computational costs.  The computational complexity of the Transformer architecture increases quadratically with image resolution, while Convolutional Neural Networks (CNN) architectures are forced to stack deeper convolutional layers to expand their receptive fields, leading to an explosive growth in computational demands.  To address these computational constraints, we leverage Mamba's linear complexity for efficiency.  However, Mamba's performance declines for small targets, primarily because small targets occupy a limited area in the image and have limited semantic information.  Accurate identification of these small targets necessitates not only Mamba's global attention capabilities but also the precise capture of fine local details.  To this end, we enhance Mamba by developing the Enhanced Small Target Detection (ESTD) module and the Convolutional Attention Residual Gate (CARG) module.  The ESTD module bolsters local attention to capture fine-grained details, while the CARG module, built upon Mamba, emphasizes spatial and channel-wise information, collectively improving the model's ability to capture distinctive representations of small targets.  Additionally, to highlight the semantic representation of small targets, we design a Mask Enhanced Pixel-level Fusion (MEPF) module for multispectral fusion, which enhances target features by effectively fusing visible and infrared multimodal information.  
\end{abstract}

\section{Introduction}
\label{sec:intro}
Target detection is an important field that has been widely utilized in urban vehicle detection~\cite{10659747}, wildlife observation~\cite{oishi2024detecting} and agricultural pest detection~\cite{wang2024vegetable}, \etc. These application scenarios often face challenges related to the low detection performance of small targets, even with the high resolution of the entire image. Furthermore, with advancements in remote sensing technology, image resolution continues to improve. However, existing CNN and Transformer networks exhibit low real-time computing efficiency when processing high-resolution images. Introducing a selective structured state space model (Mamba) with linear time complexity to target detection tasks is a promising choice to decrease the computational cost for high-resolution images. 

However, simply applying Mamba (\eg, Vmamba~\cite{liu2024vmamba} and MambaYOLO~\cite{wang2024mamba}) for small target detection shows unsatisfactory accuracy. This is primarily due to two key challenges. First, small targets occupy only a minimal percentage of the entire image, with limited semantic information available for detection. Second, the vast amount of background noise may \textit{drown out} the subtle yet crucial clues of small targets, which makes existing feature extraction networks hard to effectively capture the distinctive characteristics of small targets. Therefore, enhancing small target saliency in complex backgrounds is crucial for improving detection accuracy, which requires better networks and strategies for feature extraction and noise suppression. %Therefore, developing mechanisms to enhance the saliency of small target features amidst complex backgrounds is essential for improving detection accuracy. Addressing this challenge requires not only advanced architectures but also innovative approaches to feature representation and noise suppression.
\begin{figure}[t!]
    \begin{center}
        \includegraphics[width=0.5\textwidth]{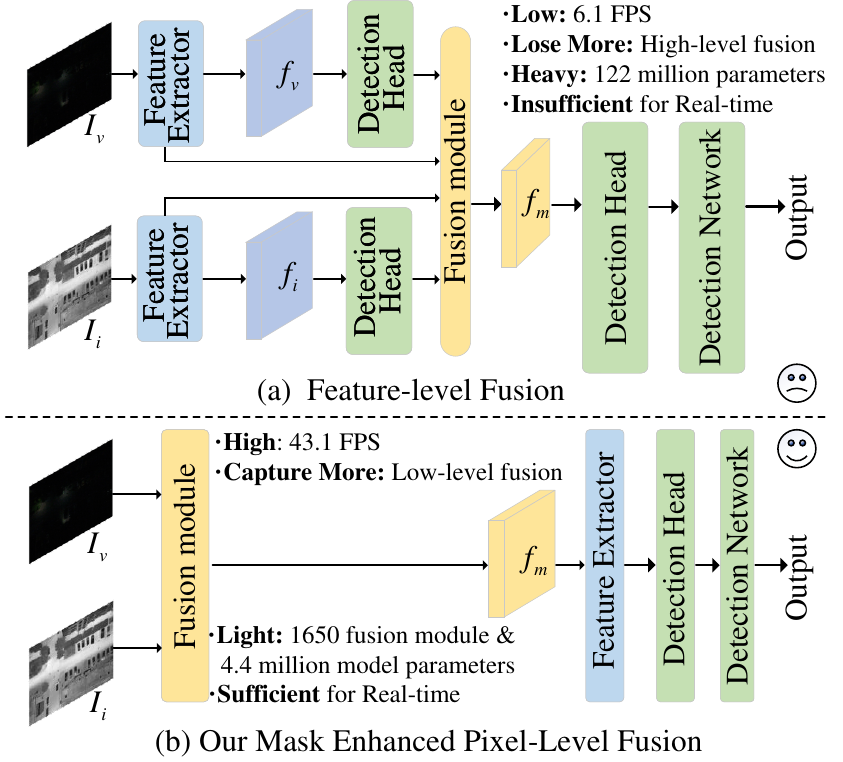}
        \caption{The comparison of previous multispectral fusion detection methods (a) and Ours (b). 
        } 
        \label{fig:MOTIVATION2}
    \end{center}
    \vspace{-1em}
\end{figure}
%%%%%%%%%%%%%%%%%%%%%%%%%%%%%%%%%%%%%%%%%%%%%%%%%%%%修改于  2024年11月15日  7:29%%%%%%%%%%%%%%%%%%%%%%%%%%%%%%%%%%%%%%%%%%%%%%%%%
% 多光谱融合检测方法可以丰富小目标特征从而有效提高小目标检测精度。融合方法主要有像素级融合\cite{10075555, Geng_2024_CVPR, Zheng_2024_CVPR}和特征级融合\cite{Ren_2024_CVPR, Yang_2024_CVPR}。特征级融合相较于像素级融合具有较多的参数量。我们主要考虑轻量的像素级融合方法。然而现有的像素级融合方法存在融合质量低的问题。不同模态的数据在进行融合时需要转换到统一的特征空间或表示形式，这会导致丢失一些原始的特征信息。如\cref{fig:MEPFMODULE}所示。To address this problem, we designed the mask generation module，能够学习更丰富的空间特征。此外，不同模态存在重复的信息，将增加计算成本，并干扰模型对关键信息的提取和理解，降低融合的效果。为此，我们设计了feature fusion module可以生成更高质量的融合图像。
Multispectral fusion detection, such as visible-infrared fusion, enhances small-target detection by leveraging complementary spectral information. Visible spectra provide texture and color details, while infrared captures thermal radiation, invariant to lighting. This synergy highlights crucial high-frequency features like edges and fine details, improving detection accuracy. Fusion methods include pixel-level~\cite{10075555, Geng_2024_CVPR, Zheng_2024_CVPR} and feature-level fusion~\cite{Ren_2024_CVPR, Yang_2024_CVPR}. 

However, existing methods face deployment challenges due to high computational costs. \eg, The Mask-guided Mamba Fusion model (MGMF)~\cite{10659747} has 122 million parameters and runs at just 6.1 FPS, limiting its suitability for mobile or real-time applications, as shown in~\cref{fig:MOTIVATION2} (a). Lightweight pixel-level fusion is a promising solution, but current approaches suffer from spatial misalignment caused by registration errors, leading to blurred or misplaced details. Moreover, aligning multi-modal features at the pixel level is difficult, exacerbating feature loss. To address these issues, we propose a mask generation module to learn aligned spatial semantics and a feature fusion module to reduce redundancy and enhance pixel-level fusion quality, as shown in~\cref{fig:MOTIVATION2} (b).
To address the challenges in small target detection, this work proposes a model based on a selective structured state space model (Mamba) and multispectral fusion. We tackle three key problems: 1) limited feature extraction and localization for small targets due to their minimal area, semantic information, and interference from complex background occlusion; 2) inefficient fusion of visible and infrared spectral information at the pixel level; and 3) high computational costs of traditional methods when processing high-resolution images. We leverage Mamba's global attention and design the Enhanced Small Target Detection (ESTD) module to enhance feature extraction, while introducing the Convolutional Attention Residual Gate (CARG) Block, built upon Mamba, to emphasize spatial and channel-wise information, reducing background noise and improving robustness. We also propose the lightweight Mask Enhanced Pixel-level Fusion (MEPF) module for efficient multispectral fusion. Additionally, we use Mamba's linear complexity to efficiently process high-resolution images. Experimental results demonstrate that we achieve high-precision performance with minimal computational cost.

Our contributions are summarized as follows:
\begin{itemize}
%1）我们为VMamba引入了增强小目标检测（ESTD）模块和卷积注意残差门（CARG）模块以有效的增强小目标特征的提取能力，缓解先前方法的漏检和误检问题。
%2）我们提出了一种用于可见光-红外小目标检测的像素级融合方法MEPF。解决了先前方法MGMF\cite{10659747}由于具有1.22亿个参数量，达不到实时部署的问题。我们MEPF的parameters仅有1650个（0.0063MB），适用于无人机等边缘设备轻量部署。
%3）在具有挑战性的基准数据集DroneVehicle\cite{9759286}上的实验证明了我们所提出的$S_{6}^{4}$-MSTD方法在可见光-红外模态融合小目标检测方面取得了新的一流性能。消融研究进一步证明了我们的方法在多模态融合和小目标检测方面的有效性。
%%%%%%%%%%%%%%%%%%%%%%%11月7日修改%%%%%%%%%%%%%%%%%%%%%%%%%%%
%针对现有视觉状态空间模型在小目标特征提取时存在特征丢失的问题，我们引入了Enhanced Small Target Detection(ESTD)模块和Convolutional Attention Residual Gate(CARG)模块以有效的增强小目标特征的捕获能力。

%针对现有融合方法存在数据对齐过程中信息丢失以及信息冗余的问题，我们引入了Mask Enhanced Pixel Level Fusion (MEPF)模块以提高融合图像质量。并且我们的MEPF的parameters仅有1650个（0.0063MB），适用于边缘设备轻量部署。

%3）在具有挑战性的基准数据集DroneVehicle\cite{9759286}和 VEDAI~\cite{Razakarivony2016Vehicle} 上的实验证明了我们所提出的$S_{6}^{4}$-MSTD方法在小目标检测方面取得了最高的mAP，在模型大小和检测帧率方面做到了即轻量又实时。消融研究进一步证明了我们的方法在多模态融合和小目标检测方面的有效性。
 
%%%%%%%%%%%%%%%%%%%%%%%11月7日修改%%%%%%%%%%%%%%%%%%%%%%%%%%
% %########################2025.3.4########################
% \item Aiming at the problem of information loss and redundancy in the process of data alignment in the existing fusion methods, we introduce the Mask Enhanced Pixel-level Fusion (MEPF) module to improve the quality of the fused image. Our MEPF has only 1650 parameters (0.0063MB), which is suitable for lightweight deployment of edge devices.
\item We propose the MEPF module for efficient multispectral fusion, which improves fusion quality with only 1650 parameters (0.0063MB), enabling lightweight deployment on edge devices.
\item The ESTD and CARG modules are proposed to improve the Mamba network, enhancing the model's capability to effectively capture visual clues of small targets. 
%########################2025.3.4########################
% \item Aiming at the problem of feature loss in feature extraction of small targets in existing visual state space models, We introduce the Enhanced Small Target Detection (ESTD) module and the Convolutional Attention Residual Gate (CARG) module to effectively enhance the ability to capture small target features.
% %########################2025.3.4########################%针对现有视觉状态空间模型中小目标特征提取存在的特征丢失问题，引入了增强小目标检测（Enhanced small Target Detection， ESTD）模块和卷积注意残差门（Convolutional Attention Residual Gate， CARG）模块，有效增强了小目标特征的捕获能力。
%We introduce the ESTD module and the CARG module for VMamba to efficiently enhance the extraction of small-target features and alleviate the leakage and false-detection problems of the previous methods.%{\color{red}[Backbone Network]}

%We propose a pixel-level fusion method MEPF for visible-infrared small target detection. Solves the problem that the previous MGMF~\cite{10659747} method could not reach real-time deployment due to having 122 million parameters. Our MEPF has only 1650 parameters, which is suitable for light deployment of edge devices \eg drones.%{\color{red}[Pixel-level fusion]}
%########################2025.3.4########################
\item Our method achieves high-precision on DroneVehicle and VEDAI datasets, balancing detection accuracy and real-time inference speed.  The experiments validate its effectiveness in multimodal fusion and small target detection.
% \item Experiments on the challenging datasets DroneVehicle~\cite{9759286} and VEDAI~\cite{Razakarivony2016Vehicle} demonstrate that our proposed $S_{6}^{4}$-MSTD method achieves the highest mAP for small target detection. In addition, it achieves both lightweight and real-time in terms of model size and detection frame rate. The ablation study further demonstrates the effectiveness of our method in multimodal fusion and small target detection.
%########################2025.3.4########################
%Experiments on the challenging benchmark dataset DroneVehicle~\cite{9759286} and VEDAI~\cite{Razakarivony2016Vehicle} demonstrate that our proposed $S_{6}^{4}$-MSTD method achieves new best-in-class performance in visible-infrared modal fusion small target detection. Ablation studies further demonstrate the effectiveness of our method for multimodal fusion and small target detection.%{\color{red}[ val DroneVehicle]}
\end{itemize}

\section{Related Work}
\label{sec:RelatedWork}

\noindent{\textbf{High-resolution image target detection.}}
As image resolution continues to increase, computational efficiency for high-resolution detection has become a focal point. CNN~\cite{7298594, 7780459, 8099726} models are inefficient for high-resolution images due to their convolution operations, which increase computation drastically with resolution. Deeper networks are often needed for global information. Vision Transformer~\cite{dosovitskiy2021an} has approached top performance in image recognition but is limited by quadratic complexity. $C\textsuperscript{2}Former-S\textsuperscript{2}ANET$~\cite{yuan2024c} (based on Transformer) can't handle resolutions above 1092$\times$1092 pixels, while DMM~\cite{zhou2024dmm} (based on Mamba) can handle 3016$\times$3016 pixels. Vision Mamba~\cite{VisionMamba} and VMamba~\cite{liu2024vmamba} use Mamba's linear complexity, showing advantages in high-resolution processing. Vision Mamba is 2.8$\times$faster and saves 86.8\% GPU memory than DeiT~\cite{hugo2021training} on 1248$\times$1248 images. Mambaout~\cite{yu2024mambaout} also highlights Mamba's potential for long sequence tasks. Traditional methods (CNN, Transformer) overlook the challenges of target detection in high-resolution images, while our approach embeds Mamba into the feature extraction backbone for efficient high-resolution image detection.

\noindent{\textbf{Small target detection.}} 
In this work, we adopt the definition of small targets from Gao \etal.~\cite{small}, where the ratio of the bounding box's width and height to those of the image is less than 0.1. Mamba-based target detection models exhibit low accuracy in detecting small targets. This issue is particularly prevalent in remote sensing, which has sparked research interest~\cite{verma2024soar,10589665,zhou2024dmm,chen2024mim}. This is because the model will focus on all areas of the entire image under global attention, but small target detection usually requires finer local details. The YOLO series is the most popular real-time target detection algorithm and widely regarded as the current standard in the field. Small target detection based on YOLO has been extensively studied~\cite{10423050, verma2024soar, 10637455, 10342786}. However, these methods still exhibit low detection performance when relying on the YOLO model based on a CNN architecture to detect small targets directly. This is because local attention, when processing small targets, loses information due to changes in the location of the small target within the local area. Existing methods often neglect balancing global and local attention in small target detection. Our approach enhances Mamba's global attention with local modules, capturing both contextual dependencies and fine-grained details. This dual mechanism ensures comprehensive feature representation, overcoming the limitations of relying solely on global attention.

 \noindent{\textbf{Multispectral fusion target detection.}}
 Multispectral fusion methods can be roughly categorized into pixel-level, feature-level, and decision-level fusion. Feature-level fusion is the fusion of extracted image features. MGMF~\cite{10659747} uses candidate regions of one modality to guide feature extraction of another modality for cross-modal feature fusion, but its performance falls short of real-time processing. Decision-level image fusion is done before getting the final detection result; this method has lower performance. Pixel-level fusion is performed on the pixels of the source image to obtain the fused image. Its advantage is that it preserves the spectral and spatial characteristics of the input image, generating a fused image that contains more information than the input images. SuperYOLO~\cite{10075555} improves target detection through pixel-level multimodal fusion. Although the method has a small number of parameters, it is expected to be improved in terms of multispectral feature fusion for small targets. Existing methods overlook the trade-off between fusion quality and computational efficiency for small targets; our approach resolves this through a lightweight fusion module that enhances feature representation while preserving real-time capability.
\section{Preliminaries}

%begin<待办1：>进行修改——把下图的内容填空到下面填空中，以及几个红色的标题也要融入到段落中【qqz over】
% 3. Preliminaries
% 在这一部分中，我们首先回顾了3.1节中的【mamba】【VMamba】，为理解【线性复杂度】问题奠定了基础。随后，我们在3.2节中介绍了【Vim】【注意引用文献】及其【检测】过程。
%在这一部分中，我们首先回顾了State Space Models (SSMs)~\cite{gu2021efficiently}为理解线性复杂度问题奠定基础。随后，我们介绍了VMamba~\cite{VMamba} 图像检测过程原理。

This section reviews State Space Models (SSM) \ie, Mamba ~\cite{gu2021efficiently} as a foundation for understanding linear complexity problems. Subsequently, we present the principles of the VMamba~\cite{liu2024vmamba} image detection process.%{\color{red}[chapter overview]}

\subsection{Mamba}
SSM maps an input sequence $x(t) \in \mathbb{R}$ to an output sequence $y(t) \in \mathbb{R}$ through a hidden state $h(t) \in \mathbb{R}^\mathtt{N}$. In the 1-D situation, discrete SSM transform sequences as linear ordinary differential equations (ODEs):
\begin{equation}
\begin{aligned}
\label{eq:lti}
h'(t) &= \mathbf{A}h(t) + \mathbf{B}x(t), \\
y(t) &= \mathbf{C}h(t).
\end{aligned}
\end{equation}
where $\mathbf{A} \in \mathbb{R}^{\mathtt{N} \times \mathtt{N}}$ is the state transition matrix, while $\mathbf{B} \in \mathbb{R}^{\mathtt{N} \times 1}$ is the input coefficient vector, and $\mathbf{C} \in \mathbb{R}^{1 \times \mathtt{N}}$ serves as the output coefficient vector.

Mamba~\cite{gu2023mamba, dao2024transformers} is an SSM-based model with selective state space. To improve expressiveness and flexibility, Mamba proposes to make $\mathbf{A}$ and $\mathbf{B}$ dynamically dependent on inputs, enabling an input-aware selective mechanism for better state space modeling. Mamba approximates this ODE by discretizing $\mathbf{A}$ and $\mathbf{B}$ with a time step parameter $\mathbf{\Delta}$ using a zero-order hold trick:
\begin{equation}
\begin{aligned}
\label{eq:zoh}
\mathbf{\overline{A}} &= \exp{(\mathbf{\Delta}\mathbf{A})}, \\
\mathbf{\overline{B}} &= (\mathbf{\Delta} \mathbf{A})^{-1}(\exp{(\mathbf{\Delta} \mathbf{A})} - \mathbf{I}) \cdot \mathbf{\Delta} \mathbf{B}.
\end{aligned}
\end{equation}
After discretization, \cref{eq:discrete_lti} is reformulated:
\begin{equation}
\begin{aligned}
\label{eq:discrete_lti}
h_k &= \mathbf{\overline{A}}h_{k-1} + \mathbf{\overline{B}}x_{k}, \\
y_k &= \mathbf{C}h_k.
\end{aligned}
\end{equation}

%这显然是一个高效的递归模型。高效率的具体原因在于有限状态，这意味着推理时间不变，训练时间为线性。
It is an efficient recursive model due to its finite state design, enabling constant-time reasoning and linear-time training. %{\color{red}[highlight the principle of linear complexity]}
%值得关注的是，Mamba具有的顺序扫描性，适用于文本但不适用于图像。因为视觉数据本身缺乏顺序排列。
It is worth noting that Mamba's sequential scanning, designed for ordered text data, is inherently unsuitable for images due to their lack of inherent order.
%It should be noted that although mamba has linear complexity, its core module S6's has sequential scanning, which applies to text but not to images. This is because visual data itself lacks sequential arrangement.%{\color{red}[highlight that the sequential scanning of S6 is not applicable to images ]}
%【mamba：①重点突出线性复杂度的原理②重点突出S6的顺序扫描不适用于图像】

%end<待办1：>进行修改——把下图的内容填空到下面填空中，以及几个红色的标题也要融入到段落中【qqz over】

%begin<待办2：>进行修改——把下图的内容填空到下面填空中，以及几个红色的标题也要融入到段落中【Vmamba】【qqz over】
%【VMamba：①重点突出SS2D解决了S6顺序扫描的问题。②突出将乘法分支消除为了提高效率，变成了网络分支和两个残差模块的结构】

%改进后 VMamba也是符合的线性复杂度的。图4的(b)展现了分辨率和flops的关系是线性的   要说明一下

% 3 . 1 .Vmamba核心思想
%vmamba
\subsection{VMamba}
\begin{figure}[t!]
    \begin{center}
        \includegraphics[width=0.46\textwidth]{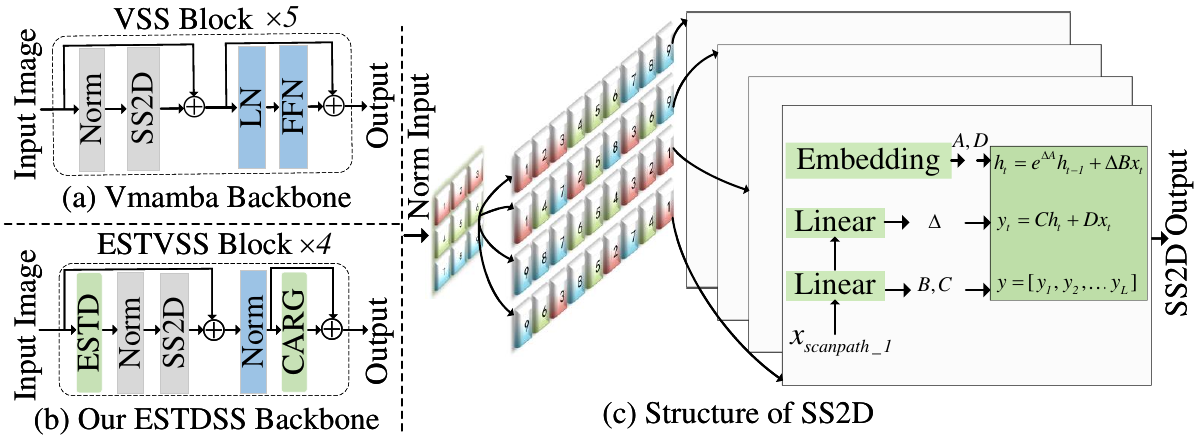}
        \caption{The comparison of Vmamba (a) and Our method (b). And structure of SS2D (c). Four different scan paths are set for traversing, and each sequence is processed independently.
        } 
        % 2d选择性扫描（SS2D）示意图。设定了四个不同的扫描路径进行遍历，每个序列分别独立处理。
        \label{fig:ss2d}
    \end{center}
    \vspace{-1em}
\end{figure}

%VMamba考虑到了视觉数据本质上是非连续的，并且包含例如局部纹理和全局结构的空间信息。其核心VSS Block模块中的SS2D结构解决了顺序扫描问题。VSS Block通过使用网络分支和两个残差模块结构而不是乘法分支，提高了计算效率。
Vmamba takes into account that visual data itself is unordered and contains spatial information about local textures and global structures. The 2D Selective Scan (SS2D) module in its core Visual State Space (VSS) Block module solves the sequential scanning problem, as shown in~\cref{fig:ss2d} (c). VSS Block improves computational efficiency by using network branches and two residual module structures instead of multiplicative branches, as shown in~\cref{fig:ss2d} (a). Our method further enhances the core structure of Mamba to address the unique characteristics of small targets, as shown in~\cref{fig:ss2d} (b).%{\color{red}[Highlighted the elimination of the multiplication branch into a structure of a network branch and two residual modules for efficiency]}

% \begin{figure}[h]
%     \begin{center}
%         \includegraphics[width=0.2\textwidth]{sec/fig/motivation2/intro25-总览备份-另一半.pdf}
%         \caption{The comparison of Vmamba (a) and Ours (b). 
%         } 
%         \label{fig:MOTIVATION25}
%     \end{center}
%     \vspace{-1em}
% \end{figure}
% \begin{figure}[h]
%     \begin{center}
%         \includegraphics[width=0.48\textwidth]{sec/fig/module/mSS2D.pdf}
%         \caption{Structure of SS2D. Four different scan paths are set for traversing, and each sequence is processed independently.
%         } 
%         % 2d选择性扫描（SS2D）示意图。设定了四个不同的扫描路径进行遍历，每个序列分别独立处理。
%         \label{fig:ss2d}
%     \end{center}
%     \vspace{-1em}
% \end{figure}

%%%%%%%%%%%%%%%%%%%%%%%%%%%%%%

% 【Vit】工作【引用vision transformer文献】提出将【图像且跨】表示为一组【XXX】【用符号和公式表示】，并使用【xxx】来检测目标。每个【token】的【xxx】参数由定义在【xxx】中的【xxx】和【xxx】确定：
% 由于【xxx】过程【xx】且【xx】，使用【xxx防范】可以有效地优化【xxx】参数。在【xxx】过程中，自适应地添加和删除【xxx】，以更好地表示【xxx】。我们将读者参见【vim或者vmamba的论文】。

%end<待办2：>进行修改——把下图的内容填空到下面填空中，以及几个红色的标题也要融入到段落中【Vmamba】【qqz over】

%begin<待办3：>进行完善【qqz over】
\section{Method}

%本节概述了我们的增强型方法的架构和具体设计要素。根据之前的分析，我们介绍了一种为无人机遥感场景量身定制的多模态融合实时小目标识别网络，如图{\color{green}[fig  XXX]}所示。该网络被命名为$S_{6}^{4}$-MSTD，由三个核心部分组成：像素级多模态融合模块、主干和包含颈部的头部。
\begin{figure*}
    \begin{center}
        \includegraphics[width=1\textwidth]{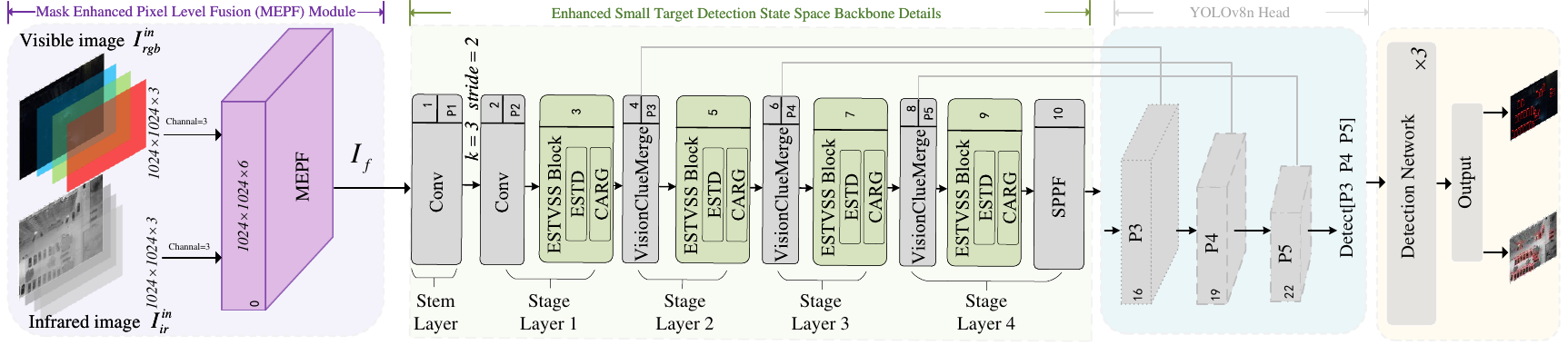}
        \caption{An overview of $S_{6}^{4}$-MSTD. It consists of four components which are the Mask Enhanced Pixel-level Fusion Module, the Enhanced Small Target Detection State Space Backbone, the YOLOv8n Head, and the Detection Network. The captured visible and infrared images are firstly fused by the MEPF module to generate the fused images $I_f$. The task of the backbone module is to extract features from the fused images. We introduce the ESTD module and CARG module to enhance Mamba's feature extraction capability specifically for small targets. After this, the head network fuses the features extracted at different stages. Finally, it is fed into the detection network to get the detection results.
        } 
        \label{fig:over}
    \end{center}
    \vspace{-1em}
\end{figure*}

This section outlines the architecture and specific design elements of our approach. We present a multimodal fusion real-time small target recognition network tailored for remote sensing scenarios: $S_{6}^{4}$-MSTD, as shown in~\cref{fig:over}. $I_{rgb}^{in}\in\mathbb{R}^{H\times W \times 3}$ is the visible modal picture, and the infrared modal picture is $I_{ir}^{in}\in\mathbb{R}^{H\times W \times 3}$. These images are taken as input to $S_{6}^{4}$-MSTD. 

% The network, named $S_{6}^{4}$-MSTD, consists of three core components: a pixel-level multimodal fusion module, a backbone, and a head containing a neck.{\color{red}[ General description of methodology, tools and techniques]}
%采集到的可见光和红外图像首先通过掩码增强像素级融合模块（MEPF）生成融合后的图像。主干模块的任务是从融合输入中提取特征。这些特征随后被送入 Head 网络。在主干部分，我们引入增强小目标检测（ESTD）模块和卷积注意残差门（CARG）模块来优化Vmamba基础。我们沿用了Yolov8n的Head网络结构。
% The acquired visible and infrared images are first passed through the Mask Enhanced Pixel Level Fusion (MEPF) module to generate the fused images. The task of the backbone module is to extract features from the fused input. These features are then fed into the Head network. In the backbone section, we introduce the Enhanced Small Target Detection (ESTD) module and the Convolutional Attention Residual Gate (CARG) module to optimise the Vmamba basis. We follow the structure of Yolov8n's Head network.{\color{red}[[ 1. Model structure (what it is) 2. Principle (why it achieves the desired goal)] }

\begin{figure}
    \begin{center}
        \includegraphics[width=0.45\textwidth]{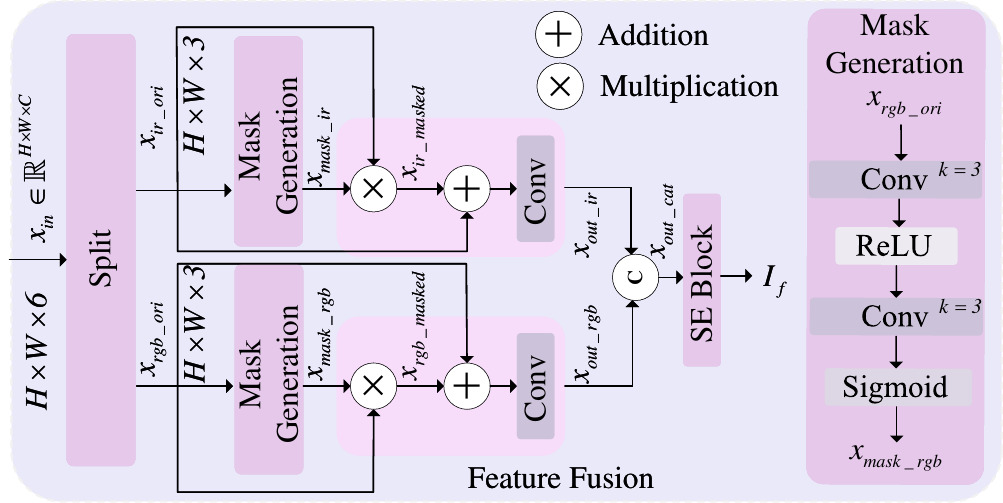}
        \caption{The architecture of the mask enhanced pixel-level fusion module consists of a split module, a mask generation module, and a feature fusion module, where $x_{in}$ is the input, $I_f$ is the pixel-level fused image, and $k=3$ indicates a convolutional kernel size of 3$\times$3. We will detail the implementation in \ref{sec:mepf}.
        } 
        % Mask Enhanced Pixel Level Fusion (MEPF) Module由输入模态信息处理模块、mask generation模块、feature fusion模块组成。其中，$x_in$为输入，$I_f$为像素级融合后的图像，$k=3$表示卷积核大小为3 $\times$ 3.我们将在\ref{sec:mepf}中详细说明具体实现。
        \label{fig:mepf}
    \end{center}
    \vspace{-1em}
\end{figure}

\subsection{Mask Enhanced Pixel-level Fusion}\label{sec:mepf}

To generate a high-quality fusion image $I_f$, we design the MEPF module.
~\cref{fig:mepf} shows our MEPF module implementation in detail. The input is $x_{in}\in\mathbb{R}^{H\times W \times 6}$. Visible modal feature $x_{rgb\_ori}\in\mathbb{R}^{H\times W \times 3}$ and infrared modal feature $x_{ir\_ori}\in\mathbb{R}^{H\times W \times 3}$ are obtained after split module.  
%$x_{left\_ori}$和$x_{right\_ori}$将被送入mask generation模块生成初步的mask，如公式\cite{eq:MEPF1}所示。

%$x_{left\_ori}$ and $x_{right\_ori}$ will be fed into the mask generation module to generate preliminary masks~\cref{eq:MEPF1}.
%mask generation模块的详细
%其中,$\sigma_R, \sigma_S$分别为 ReLU 和 Sigmoid 函数。
% \begin{equation}
% \begin{aligned}
% x_{mask\_left} &= \sigma_{Sig}(C_3(\sigma_{ReLU}(C_3(x_{left\_ori})))),\\
% x_{mask\_right} &= \sigma_{Sig}(C_3(\sigma_{ReLU}(C_3(x_{right\_ori}))))          
% \end{aligned}
% \label{eq:MEPF1}
% \end{equation}
% where,$\sigma_{ReLU}, \sigma_{Sig}$ are ReLU and Sigmoid functions, $C_3$ is ${Conv}_{3\times 3}$.
%考虑到不同模态的数据在进行融合时需要转换到统一的特征空间或表示形式，这会导致丢失一些原始的空间结构信息或深度信息。针对此问题，我们设计了如 \cref{eq:MEPF1}中所示的mask generation module to generate the initial mask $x_{mask\_rgb}$ and $x_{mask\_ir}$.
When the data of different modes are fused, they need to be transformed into a unified feature space or representation, which misses some spatial structure information or depth information. To address this problem, we design the mask generation module shown in~\cref{eq:MEPF1} to generate the initial mask $x_{mask\_rgb}$ and $x_{mask\_ir}$.
\begin{equation}
x_{mask\_rgb} = \sigma_{Sig}(C_3({ReLU}(C_3(x_{rgb\_ori}))))
\label{eq:MEPF1}
\end{equation}
where ${ReLU}$ is ReLU activation~\cite{bai2022relu}, $\sigma_{Sig}$ is Sigmoid functions~\cite{elfwing2018sigmoid} and $C_3$ is ${Conv}_{3\times3}$~\cite{Szegedy_2015_CVPR}.
%x_{mask\_rgb} = \sigma_{Sig}(C_3(\sigma_{ReLU}(C_3(x_{rgb\_ori}))))
%where, $\sigma_{ReLU}$ is ReLU activation~\cite{bai2022relu}, $\sigma_{Sig}$ is Sigmoid functions~\cite{elfwing2018sigmoid} and $C_3$ is convolutions with spatial kernel size of 3$\times$3.

%生成的$x_{mask\_left}$和$x_{mask\_right}$应用到特征融合网络。如公式\cite{eq:MEPF2}所示。
%每种模态数据本身存在一定的冗余信息。例如，图像中的背景信息、重复的纹理等。在多模态融合时如果没有对这些冗余信息进行有效的筛选和去除，会增加融合后特征的维度和计算复杂度。针对此问题，我们设计了如 \cref{eq:MEPF2}中所示的feature fusion module。
Each modality's data itself has some redundant information, \eg, background information and repetitive textures. If the redundant information is not screened and removed effectively in multi-modal fusion, the computational complexity of the fused feature would increase. To address this, we design a feature fusion module (shown in~\cref{eq:MEPF2}) that selectively filters redundant information while preserving critical features.
\begin{equation}
x_{out\_rgb} = C_3(x_{rgb\_ori} + x_{rgb\_ori} \otimes x_{mask\_rgb})
\label{eq:MEPF2}
\end{equation}
where $\otimes$ is the element-wise multiplication.

%之后，将两种模态的特征进行最后的融合，从而生成$I_f$。
% Afterwards, the features of the two modalities are finally fused to generate $I_f$.~\cref{eq:MEPF3}.

%不同模态之间存在一些重复或相似的信息。如果简单地将这两种模态的数据进行拼接或融合，就会引入大量的冗余信息。这些冗余信息不仅会增加计算成本，还可能会干扰模型对关键信息的提取和理解，降低融合的效果。针对此问题，我们设计了如 \cref{eq:4}中所示的方法。即Along the spatial dimension, feature compression is executed subsequently, converting each two-dimensional feature channel from multiple modalities into a modal factor $M \in \mathbb{R}^{(C_1+C_2)\times 1 \times 1}$, \ie
%%%%%%%%%%%%%%%%%%%%%%%%%%%%%%%%%%%%%%%%%%%%%%%%%%%%%%%%%%%%%%%%%%%%%%%%%%%%
% There exists a certain degree of duplication or analogous information among the different modes. If the data of these two modes are simply spliced together or fused, a substantial amount of redundant information will be introduced. Such redundant information will not merely augment the computational cost but disrupt the model's extraction and comprehension of crucial information, thereby leading to a diminished fusion effect. To address this, we devise the method depicted in~\cref{eq:4}.  \ie along the spatial dimension, feature compression is executed subsequently, converting each two-dimensional feature channel from multiple modalities into a modal factor $M \in \mathbb{R}^{(C_1+C_2)\times 1 \times 1}$.
%%%%%%%%%%%%%%%%%%%%%%%%%%%%%%%%%%%%%%%%%%%%%%%%%%%%%%%%%%%%%%%%%%%%%%%
Furthermore, we propose a spatial feature compression method (depicted in~\cref{eq:4}), which compresses multi-modal 2D feature channels into a compact modal factor $M \in \mathbb{R}^{(C_1+C_2)\times 1 \times 1}$, effectively reducing redundancy while maintaining essential information.
\begin{equation}\label{eq:4}
\begin{aligned}
M &= \sigma_{Sig}(FC({ReLU}(FC(M')))),\\
M'&= Avg(Cat[x_{out\_rgb}, x_{out\_ir}])
\end{aligned}
\end{equation}
where $FC$ is the fully connected layer~\cite{krizhevsky2012imagenet}, $Avg$ is the global average pooling~\cite{lin2013network}. The factor $M$ captures a global receptive field, with its output dimension matching the input feature channels. This enables layers near the input to gain broader context awareness, enhancing their understanding of contextual information. The result is shown in \cref{eq:MEPF3}.
\begin{equation}
I_f = M \cdot (Cat[x_{out\_rgb}, x_{out\_ir}])
\label{eq:MEPF3}
\end{equation}

%%%%%%%%%%%%%%%%%%%%%%%%%%%%11月9日%%%%%%%%%%%%%%%%%%%%%%%%%%%%%
%在图XX中的baseline中使用1×1的卷积生成掩码。简单的1x1卷积不足以捕捉复杂的空间关系，从而影响mask的质量。相比之下，我们的MEPF使用两个3x3卷积层和一个ReLU激活函数来生成mask，能够学习更丰富的空间特征。在mask应用方式上，baseline将mask与0.5倍缩放后的特征相乘。这种缩放会导致信息丢失。相比之下，我们直接地应用mask到原始特征图上，并且使用Sigmoid函数将mask值限制在0到1之间。一方面，我们没有额外的缩放步骤可以更好利用特征的原始信息。另一方面，可以更精细地选择有效特征。我们的方法可以生成更高质量的融合图像。
~\cref {fig:MEPFMODULE} compares the feature map visualizations of our MEPF module and baseline.  The baseline~\cite{10075555} uses a 1$\times$1 convolution to generate a mask. The simple 1$\times$1 convolution is unable to capture the complex spatial relationships. In contrast, our MEPF uses two 3$\times$3 convolution layers and a ReLU activation function to generate masks, capable of learning small target-matched spatial features. In the mask application mode, the baseline multiplies the mask with the features after 0.5 times scaling. This scaling results in information loss. In contrast, we apply the mask directly to the original feature map and use the Sigmoid function to rescale the mask value to $[0,1]$. We avoid additional scaling operations, thereby making better use of the original features.

%%%%%%%%%%%%%%%%%%%%%%11月9日%%%%%%%%%%%%%%%%%%%%%%%%%%%%%%%%%%%%%%%
% \begin{equation}
% x_{mask\_left} = MG(x_{left\_ori}),  
% \\
% x_{mask\_right} = MG(x_{right\_ori})
%   \label{eq:MEPF1}
% \end{equation}

% XXX.{\color{red}[Propose a solution method]}

% XXX.{\color{red}[Analyze the benefits of the method] [Analyze the benefits of the method]}

% XXX.{\color{red}[Improvement of the second module in this large module, introduction of the improvement principle and advantages]}

%end<待办4：>进行修改——把下图的内容填空到下面填空中，以及几个红色的标题也要融入到段落中【qqz over】

%begin<待办5：>进行修改——把下图的内容填空到下面填空中，以及几个红色的标题也要融入到段落中【创新点2—— 引入增强小目标检测（ESTD）模块和卷积注意残差门（CARG）2个创新】【待验证是否真的有效】【qqz over】

% 为了克服这些挑战，我们对原始的【mambayolo】【引用参考文献】模型进行了两次修改。特别地，我们引入了【创新点2第一小点】，将【xxx】限制在训练图像确定的最大【xxx】以下，从而消除了【xxx】时的【xxx】。此外，我们证明了用【创新点2的第二小点】代替【xxx】，它近似【xxxx】过程中固有的【xxxxx】，有效地减轻了【xx】问题。结合起来，【MM-ViMY】可以在不同的【xxx】下实现【xxxx】。我们详细讨论了【创新点2第一小点】和【创新点2的第二小点】。

\subsection{Enhanced Small Target Visual State Space}
The design of the backbone is based on two main considerations. On one hand, the backbone should achieve high performance in processing high-resolution images.  On the other hand, it should be able to better capture the feature information of small targets.

To address the problem that the computational complexity of the Transformer architecture grows quadratically with the increase of image resolution, whereas CNN requires deeper structures to handle high-resolution images, we choose to introduce the selective structured state space model (Mamba) with linear time complexity into the visual backbone to solve the computational constraints of high-resolution image detection.

\begin{figure}
    \begin{center}
        \includegraphics[width=0.5\textwidth]{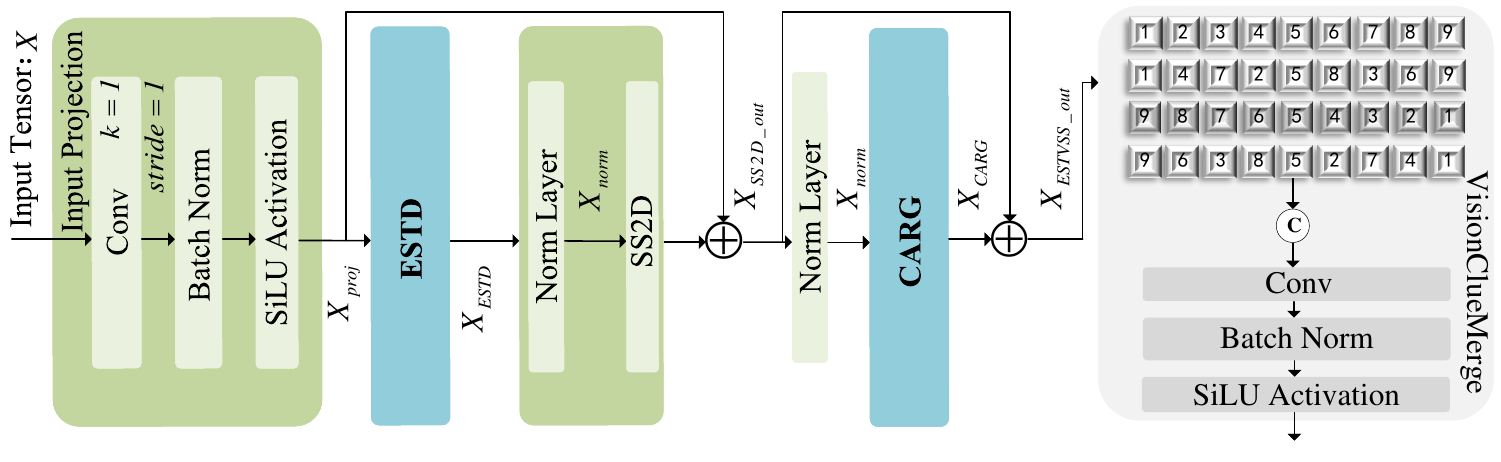}
        \caption{Structure of ESTVSS Block and VisionClueMerge. MEPF module generates $I_{f}$. After that, we use ESTVSS Block for feature extraction. VisionClueMerge is used to merge and stitch tensors. The ESTD module and the CARG module are used to enhance the detection of small targets.
        } 
        % ESTVSS Block和VisionClueMerge的详细结构图。像素级融合(MEFP)生成的[XXIfXXX]经过Stem Layer处理后送入到Stage Layer 1，我们利用ESTVSS Block进行特征提取，VisionClueMerge用于合并和拼接张量。ESTD模块和CARG模块用于增强小目标的检测能力。。
        \label{fig:module1}
        \vspace{-12pt}
    \end{center}
\end{figure}

However, Mamba's performance declines for small targets due to its reliance on long-range dependencies, which may weaken local feature representation. In high-resolution images, small targets occupy a minimal proportion, resulting in weak features that are easily overwhelmed by complex background information. To address this, we design the Enhanced Small Target Visual State Space (ESTVSS) Block (as shown in~\cref{fig:module1}), a tailored modification to the Mamba architecture for small target detection. The ESTD block enhances local attention, capturing fine-grained details that global attention often misses. The CARG block further strengthens spatial and channel attention, effectively distinguishing small targets from complex backgrounds. Together, these modifications enable the ESTVSS Block to overcome the limitations of Mamba in small target scenarios.
%该模块在mamba架构的基础上，通过引入ESTD Block和CARG Block能够捕捉更精细的局部细节。

%%%%%%%%%%%%%%%%%%%%%%%%%%%%%%%%%%%%%11月14日 11:37%%%%%%%%%%%%%%%%%%%%%%%%%%%%%%%%%%%%%%%%%%%%%

%此外，现有方法在小目标特征提取过程中会loss一些特征，因此我们设计了ESTVSS Block。该模块在mamba架构的基础上，通过ESTD Block和CARG Block能够capture更全面的信息。
% In addition, existing methods can not capture features well in the process of small target feature extraction. Therefore, we design the Enhanced Small Target Visual State Space (ESTVSS) Block (as shown in~\cref{fig:module1}). Based on Mamba architecture, this module can capture more comprehensive information through ESTD Block and CARG Block. 

%输入张量$X$经过Input Projection module处理得到$X_{proj}$。
The input tensor $X$ is processed by the Input Projection module to obtain $X_{proj}$~\cref{eq:ESTD1}.
\begin{equation}
X_{proj} = {SiLU}(BN(C_1(X)))
\label{eq:ESTD1}
\end{equation}
where $C_1$ is ${Conv}_{1\times 1}$~\cite{Szegedy_2015_CVPR}. ${SiLU}$ is SiLU activation~\cite{elfwing2018sigmoid}. BN is Batch normalization layer~\cite{ioffe2015batch}.

\begin{figure}
    \begin{center}
        \includegraphics[width=0.47\textwidth]{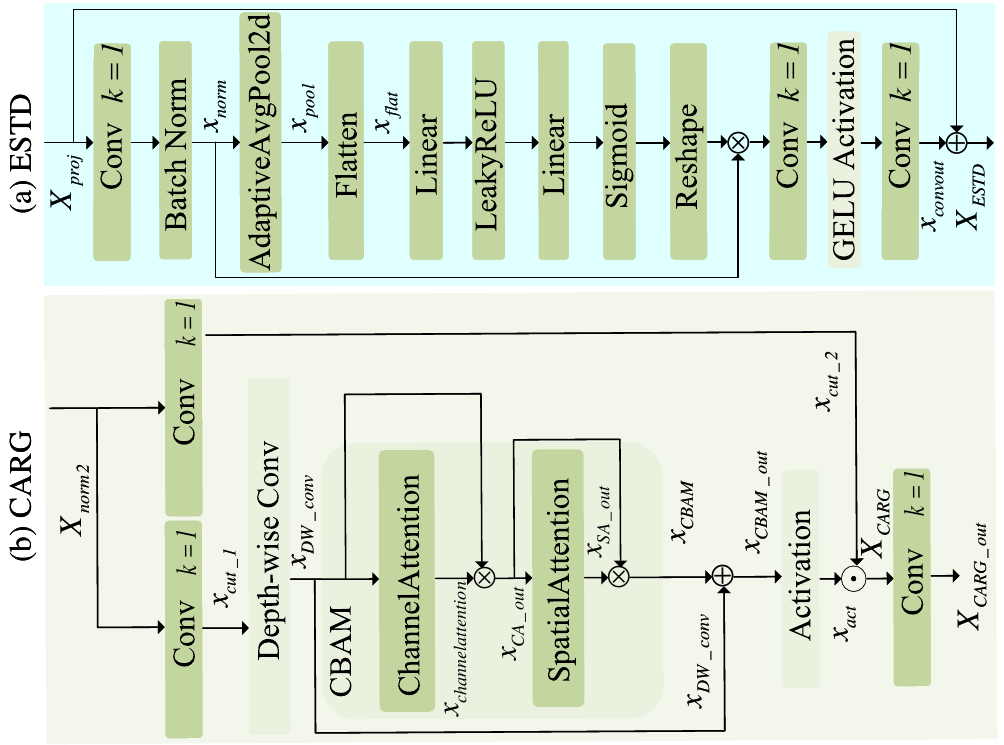}
        \caption{Structure of ESTD block and CARG block.
        } 
        % xxx。
        \label{fig:module2}
    \end{center}
    \vspace{-1em}
\end{figure}

\noindent{\textbf{Enhanced Small Target Detection Module.}}
% The size of the receptive field significantly impacts the accuracy of small target detection. To enhance the precision of detecting small targets, it is imperative to leverage both global attention mechanisms to aggregate information from a holistic image perspective and local attention mechanisms to precisely capture the intricate textures and details of the target. Depending on the global attention for feature extraction, vital details such as the texture of a small target will be missed due to a lack of targeted focus. Conversely, using only local attention mechanisms in complex and variable background environments can lead to detection performance being constrained by local interfering information. \eg When local regions exhibit textures similar to small targets, background textures will be mistakenly identified as target features. Global attention mechanisms excel at capturing contextual information from the entire image scene, which can be utilized in conjunction with this scene information to aid in determining the specific location and category of small targets. Local attention mechanisms specialize in capturing more refined local features. Their combination can significantly improve the detection accuracy of small targets. Therefore, we introduce ESTD Block into Mamba architecture to optimize the size of the receptive field by enhancing local attention and making it more suitable for small target detection. As shown in~\cref{fig:module2} (a) and~\cref{eq:ESTD2}.
The receptive field size crucially influences small target detection accuracy. Relying solely on global attention risks missing critical small target features due to insufficient focus, while using only local attention in complex backgrounds leads to interference from similar textures. Global attention aids in target localization through contextual understanding, whereas local attention excels at extracting detailed features.     Combining global attention, which captures overall image context, with local attention, focused on fine-grained details, is beneficial for precision. Therefore, we introduce the ESTD Block, a tailored modification to the Mamba architecture, specifically designed to optimize the receptive field for small target detection by enhancing local attention. As shown in~\cref{fig:module2} (a) and~\cref{eq:ESTD2}, this modification addresses the unique challenges of small targets, ensuring precise feature extraction in complex scenarios.
\begin{equation}
X_{ESTD} = C_1({GELU}(C_1(SE(BN(C_1(X_{proj}))))))
\label{eq:ESTD2}
\end{equation}
where ${GELU}$ is GELU activation~\cite{lee2023gelu}. SE is Squeeze-and-Excitation block~\cite{hu2018squeeze}. 
%SE is Squeeze-and-Excitation block~\cite{hu2018squeeze}. BN is Batch normalization layer~\cite{ioffe2015batch}.
%%%%%%%%%%%%%%%%%%%%%%%%%%%%%%%%%%%%%11月14日 11:37%%%%%%%%%%%%%%%%%%%%%%%%%%%%%%%%%%%%%%%%%%%%%
%【插入图ESTD module】

\noindent{\textbf{Convolutional Attention Residual Gate Module.}}
%我们聚焦于研究导致小目标检测准确度低的两个主要原因。即现有方法难以捕捉微弱的形状、颜色、纹理等典型特征以及未能处理好复杂背景遮挡问题。我们通过设计CARG module增强模型的通道和空间注意力。As shown in \cref{fig:module2} (b) and \cref{eq:}.
% We focus on two main reasons for the low accuracy of small target detection. \ie The first is that existing methods have difficulty capturing faint typical features \eg shape, color, texture, \etc. The second is the failure to deal with complex background occlusion. We enhance the channel and spatial attention of the model by designing the CARG module. As shown in~\cref{fig:module2} (b).
%%%%%%%%%%%%%%%%%%%%%%%%%%%%%%%%%%%%%11月14日 11:37%%%%%%%%%%%%%%%%%%%%%%%%%%%%%%%%%%%%%%%%%%%%%
%我们聚焦于研究导致小目标检测准确度低的两个主要原因。即现有方法难以捕捉微弱的形状、颜色、纹理等典型特征以及未能处理好复杂背景遮挡问题。为了解决以上问题，我们通过设计CARG module增强模型的通道和空间注意力。As shown in \cref{fig:module2} (b) and \cref{eq:}.
% The existing approach has two main limitations. 1) It is difficult to capture faint typical features \eg shape, color, texture, \etc. 2) It is not able to deal with the complex background occlusion problem, which is the main reason for the low accuracy of small target detection. To overcome these challenges, we introduce the Convolutional Attention Residual Gate (CARG) module, which enhances the Mamba model's channel and spatial attention capabilities, as illustrated in~\cref{fig:module2} (b). This tailored modification enables more robust feature extraction in complex scenarios, specifically addressing the unique challenges of small target detection.
The existing approach faces two primary limitations: 1) difficulty in capturing subtle yet critical small target features, such as shape, color, and texture; and 2) inability to handle complex background occlusion, which significantly reduces detection accuracy. To address these challenges, we introduce the CARG module, a tailored enhancement to the Mamba architecture designed specifically for small target detection. The CARG module augments the model's channel and spatial attention capabilities, as illustrated in~\cref{fig:module2} (b), enabling robust feature extraction in complex scenarios.
The CARG module's channel attention mechanism identifies and emphasizes channels containing critical semantic information, such as shape and color, while suppressing irrelevant background channels.  This selective weighting helps the model focus more on key small target features, reducing the impact of complex backgrounds.  Complementing this, the spatial attention component enhances the model's ability to distinguish small targets from occluding backgrounds.  Together, these mechanisms enable the CARG module to effectively address the unique challenges of small target detection in high-resolution imagery, making it a pivotal enhancement to the Mamba architecture.
\begin{equation}
x_{channelattention} = \sigma_{Sig}(x_{out_{avgpool}} \otimes x_{out_{maxpool}})
\label{eq:CARG8}
\end{equation}
\begin{equation}
x_{out_{avgpool}} = C_1({ReLU}(C_1(f_{AP}(x_{DW_{conv}})))
\label{eq:CARG6}
\end{equation}
\begin{equation}
x_{out_{maxpool}} = C_1({ReLU}(C_1(f_{MP}(x_{DW_{conv}})))
\label{eq:CARG7}
\end{equation}
where $f_{MP}$ is 2D adaptive max pooling~\cite{sun2023spatially}. $f_{AP}$ is 2D adaptive average pooling~\cite{yao2024deco}. $x_{DW_{conv}}$ is the feature processed by Depth-wise separable convolution~\cite{chollet2017xception}.

%Spatial Attention的主要作用是关注小目标空间位置信息和降低周边背景区域权重。小目标在图像中所占的空间区域通常很小，容易被周围的背景所掩盖。Spatial Attention机制可以通过对空间位置的分析，从而识别出小目标所在的具体空间区域，并赋予该区域较高的权重，使得模型在后续处理过程中能够将更多的注意力集中在小目标所在的空间位置上。同时会降低周边背景区域的权重，从而将模型的注意力从背景区域转移到小目标区域，排除了背景对小目标检测的干扰，使得模型能够更纯粹地处理小目标的相关信息。
% The main function of Spatial Attention is to pay attention to the location information of a small target space and reduce the weight of surrounding background regions. The area of space occupied by small targets in the image is usually small and easily obscured by the surrounding background. The Spatial Attention mechanism can identify the specific spatial region where the small target is located through the analysis of the spatial location, and assign a higher weight to the region, such that the model can focus more attention on the spatial location of the small target in the subsequent processing. Meanwhile, the weight of the surrounding background region is reduced, thereby shifting the model's attention from the background region to the small target region, eliminating the interference of the background on the small target detection, and enabling the model to process the relevant information of the small target.
The primary function of Spatial Attention is to enhance small target localization by suppressing background interference. The area occupied by small targets in the image is usually small and they are easily obscured by the surrounding background. The Spatial Attention mechanism can identify the specific spatial region where a small target is located through the analysis of spatial locations, and assign a higher weight to this region so that the model can focus more attention on the spatial location of the small target in subsequent processing. Meanwhile, the weight of the surrounding background region is reduced, thereby shifting the model's attention from the background region to the small - target region, reducing the interference of the background on small target detection, and enabling the model to process the relevant information of small targets more accurately. 
\begin{equation}
x_{mean} = ChannelMean(x_{CA_{out}})
\label{eq:CARG9}
\end{equation}
\begin{equation}
x_{max} = ChannelMax(x_{CA_{out}})
\label{eq:CARG10}
\end{equation}
\begin{equation}
x_{spatialattention}   =  \sigma_{Sig}(C_1(Cat[x_{mean},x_{max}]))
\label{eq:CARG11}
\end{equation}
where $x_{CA_{out}}$ is shown in~\cref{fig:module2} (b).

\section{Experiments}

% 我们首先介绍 【MM-ViMY】的实现细节。然后，我们评估了它在具有挑战性的 【dronevehicle】数据集【引用数据集所在的文献】上的性能。最后，我们讨论了我们方法的局限性。
% 我们首先介绍$S_{6}^{4}$-MSTD的实现细节。然后，我们评估了它在具有挑战性的 DroneVehicle数据集\cite{9759286}上的性能。最后，我们讨论了我们方法的局限性。
We first present the implementation details of $S_{6}^{4}$-MSTD. Then, we evaluate its performance on the challenging DroneVehicle~\cite{9759286} and VEDAI~\cite{Razakarivony2016Vehicle} datasets. Finally, we discuss the limitations of our approach.%{\color{red}[chapter overview]}

%end<待办1：>填写下面的空【qqz over】

%begin<待办2：>填写下面的空，并且把红色标题融进去【qqz over】
% 我们的方法建立在流行的开源 【yolov8】 代码库 【引用github代码库】上。按照【引用github代码库】 ，我们在所有场景中训练模型进行 【300轮】 迭代，并使用相同的损失函数、优化器和超参数【包括要说明学习率/优化器类型/学习策略选的哪种】。我们选择【创新点2和3】的【某个参数】为 【例如：0.1】，近似于【解释选择为0.1的原因xx】，选择 【创新点1】的【某个参数】为 【例如：0.2】，以便【xxx】。【实验中详细的参数设置】【以及公式中的取值】
% \subsection{Experimental Settings}

%% A. 实施细节 
%1) 数据集： DroneVehicle\cite{9759286}数据集是目前跨模态车辆小目标检测领域规模最大的基准数据集。该数据集包括56878张图像，即28 439 对可见红外图像。涉及五种类型的车辆，包括轿车、公共汽车、卡车、货车和货车。其中，所有收集到的图像分为三组：训练集、验证集和测试集，分别由 17 990、1469 和 8980 幅图像组成。图像大小为 840 × 712 像素。每张图像都是由装有摄像头的无人机在不同场景和不同光照条件下拍摄的，从而确保了 DroneVehicle 数据集的多样性和复杂性。
%2) 设置： The experiments are carried out on a single NVIDIA RTX A100 GPU. We implement our algorithm with Pytorch toolbox and SGD optimizer with a momentum of 0.937 and a weight decay of 0.0005. The initial learning rate is set to 0.01 and eventually reduced to 0.01. The batch size is 8. The training epoch is set to 300. We use the ground truth of the IR images as the training label, owning to the more comprehensive target annotations available in the IR modality.
\subsection{Datasets}
%\subsubsection*{Datasets}
% \noindent \textbf{nuScenes dataset}~\cite{caesar2020nuscenes}
\noindent \textbf{DroneVehicle dataset}~\cite{9759286} is currently the largest benchmark dataset in the field of cross-modal vehicle small target detection. The dataset consists of 56,878 images, \ie, 28,439 pairs of visible infrared images. Five types of vehicles are involved, including cars, buses, trucks, vans, and lorries. Among them, the training set, validation set, and test set consist of 17,990, 1,469, and 8,980 images. The image size is 840$\times$712 pixels. Each image is captured by a camera-equipped drone in different scenes and under different lighting conditions, with a large number of nighttime scenes included, thus ensuring the diversity and complexity of the DroneVehicle dataset.

\noindent \textbf{VEDAI dataset}~\cite{Razakarivony2016Vehicle} is designed for vehicle detection in high-resolution aerial imagery, including complex background environments, \eg, grasslands, highways, mountains, and urban landscapes. The dataset consists of 1246 pairs of RGB and infrared images with resolutions of 1024$\times$1024 and 512$\times$512 pixels. According to~\cite{Razakarivony2016Vehicle}, small object detection achieves only 10\% to 20\% recall at 0.01 FPPI, with most false positives from background areas. This highlights the challenge of detecting small objects in complex scenes. In our experiments, we use the 1024$\times$1024 version to evaluate our method's robustness.

% Vehicle Detection in Aerial Imagery (VEDAI) dataset~\cite{Razakarivony2016Vehicle}, which is derived from a subset of the expansive Utah Automated Geographic Reference Center (AGRC) dataset. The AGRC dataset comprises images captured from a uniform altitude, each spanning approximately 16,000 $\times$ 16,000 pixels with a pixel resolution of roughly 12.5 cm $\times$ 12.5 cm. The images in the VEDAI dataset are available in two modalities: RGB and IR, representing the same scenes. The VEDAI dataset comprises 1,246 images that spotlight a variety of settings, including grasslands, highways, mountainous regions, and urban landscapes. These images are resized to either 1024 $\times$ 1024 or 512 $\times$ 512 pixels for analysis. The objective of the dataset is to identify 11 distinct classes of vehicles, encompassing categories \eg cars, pickups, campers, and trucks.

\begin{figure}
    \begin{center}
        \includegraphics[width=0.4\textwidth]{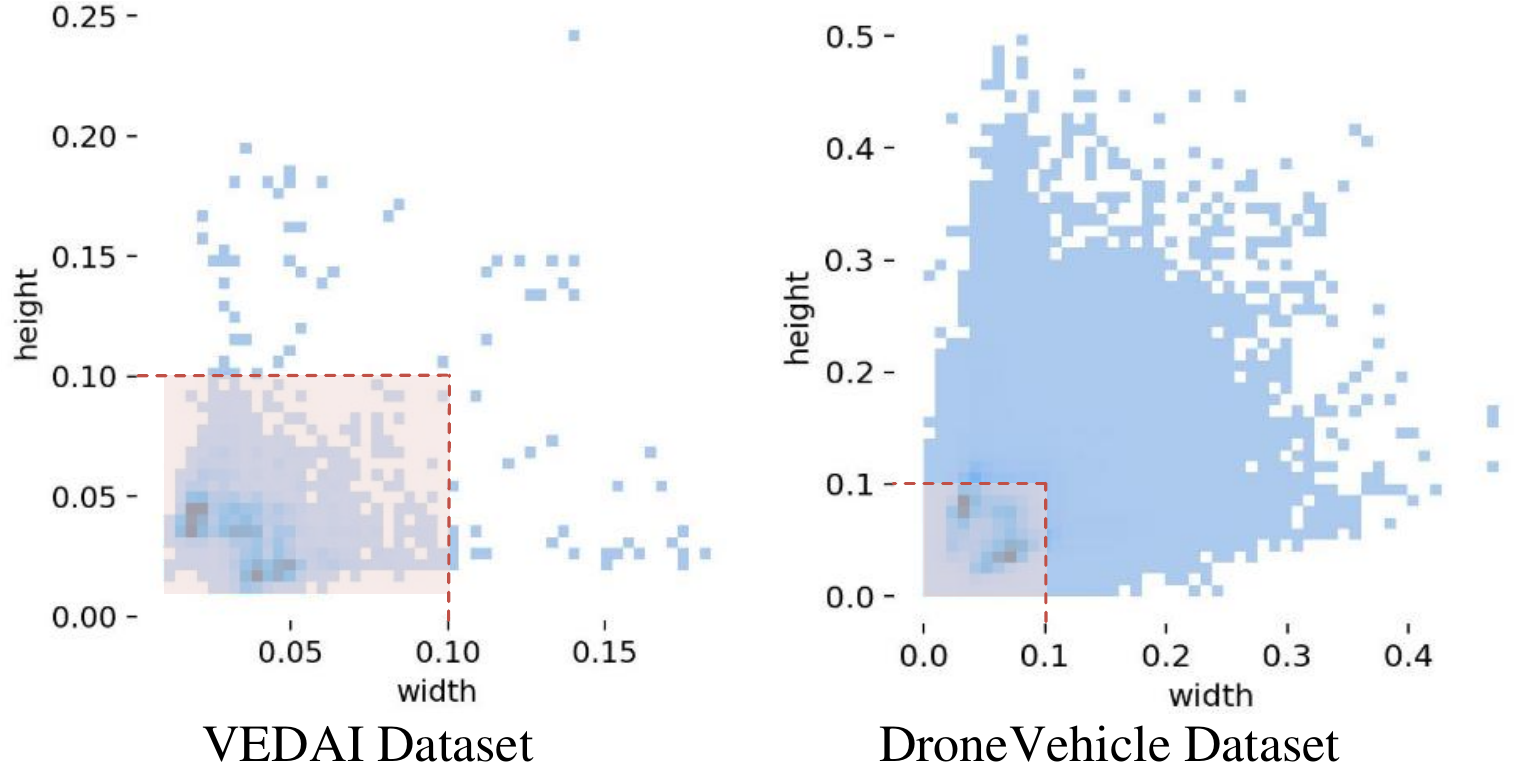}
        \caption{Distribution of target sizes in the datasets. The majority of targets in the VEDAI dataset~\cite{Razakarivony2016Vehicle} meet the definition of small targets (with a size ratio below 0.1)~\cite{small}, while only a small portion of targets in the DroneVehicle dataset~\cite{9759286} satisfy this criterion.
        } 
        % 两个数据集的目标尺寸分布对比。
        \label{fig:size}
    \end{center}
     \vspace{-1em}
\end{figure}

\subsection{Training details}
The experiments are carried out on a single NVIDIA RTX A100 GPU with 80 GB memory. We implement our algorithm with PyTorch and SGD optimizer~\cite{sutskever2013importance} with a momentum of 0.937 and a weight decay of 0.0005. The learning rate is set to 0.01. The batch size is 8. The training epochs are set to 300. We use the ground truth of the IR images as the training label, owing to the more comprehensive target annotations available in the IR modality. In addition, to ensure that the model is lightweight, we refer to the model parameter settings of YOLOv8n~\cite{yolov8_ultralytics}, \ie, the model's depth is set to 0.33 and the model's width is set to 0.25.
%此外，为了保证模型是轻量的，我们参考了YOLOv8n的模型参数设置，\ie 模型的depth设置为0.33，模型的width设置为0.25。

\subsection{Assessment Indicators}
Mean Average Precision (mAP) is utilized as a precision metric to assess the performance of the various methods. Additionally, Frames Per Second (FPS) and parameter sizes serve as quantifiable measures to evaluate the real-time capability and computational expenditure of the model.

\subsection{Results Comparisons}
% XXX。{\color{红色}[强调验证有效性][强调相同的条件][]}
%XXX.{\color{red}[Emphasize to verify the validity][Emphasize the same conditions][]}
%！！！！！！！！！！！！！！！！！！！以及   ①可以从VMamba需要更多的参数量下手吐槽，最少的也要30M  ，最大的要89M  ②VMamba只在COCO数据集上进行了目标检测任务验证，分辨率为224×224并不属于高分辨率，而我们进行了1024×1024的验证，并且有良好的效果 
% XXX。{\color{green}[[实验结果对比]Insert Table 4][[Insert Table 4]原始模型与改进模型参数对比][参数GFLOPs mAP (\%)]}

% %DMM是mamba，DPDETR和36是transformer，其余是CNN。可以很明显的看出mamba和transformer超出CNN很多

%XXX.{\color{green}[[Comparison of experimental results] Insert Table 4][[Insert Table 4 Comparison of parameters between the original model and the improved model] [Parameters GFLOPs mAP (\%)]}
%%%%%%%%%%%%%%%%%%%%%%%%%%%%%%%%%%%%%%%%%%%%%%%%%%%%%%%%%%%%%%%%%%%%%%%%%%%%%%%%%%%%%%%%%%%%%%%

%begin<待办4：>【最好是再加一个数据集VEDAI进行方法验证】【qqz over】

% 我们在 【VEDAI】数据集【数据集的引文】上进一步评估了我们的方法，遵循广泛使用的设置，其中模型以相同的比例进行训练和测试，如表 4 所示，在这个具有挑战性的基准测试中，我们的方法与 【XX论文的结果】 和 【XX论文的结果】 相当，性能没有任何下降。这证实了我们的方法处理各种设置的有效性。
% 我们设置相同的参数在VEDAI数据集\cite{Razakarivony2016Vehicle}上进行训练和测试，进一步评估了我们的方法。实验结果表明，在这个具有挑战性的基准测试中，我们的方法达到了81.2\%mAP，为最先进的性能。
%end<待办4：>【最好是再加一个数据集VEDAI进行方法验证】【qqz over】
% To compare the detection accuracy of small targets, we evaluate the DroneVehicle dataset, which includes long-wave infrared and visible image pairs with 840$\times$712 pixels. And the VEDAI dataset, which includes near-infrared and visible image pairs with 1024$\times$1024 pixels.
To evaluate the detection accuracy of small targets, we use the VEDAI dataset~\cite{Razakarivony2016Vehicle}, which contains mostly small targets (size ratio below 0.1)~\cite{small}. Additionally, to validate the applicability of our method in scenarios with mixed small and large targets, we evaluate the DroneVehicle dataset~\cite{9759286}.
%为了评估小目标的检测精度，我们使用VEDAI数据集\cite{Razakarivony2016Vehicle}，该数据集主要包含小目标（尺寸比小于0.1）\cite{small}。此外，为了验证我们的方法在大小目标混合场景中的适用性，我们评估了DroneVehicle数据集\cite{9759286}。
The experimental results on the VEDAI dataset~\cite{Razakarivony2016Vehicle} show that our method achieves 81.2\% mAP, a state-of-the-art performance in this challenging benchmark test, as shown in~\cref{tbl:table2}.

\begin{table}[ht]
    \centering
     \renewcommand\arraystretch{1.1}
     \resizebox{\columnwidth}{!}{ % Fit to single-column width
    %\scalebox{0.95}{
    \begin{tabular}{c|cccccccc|c}
        \toprule
         Method & \rotatebox{90}{Car} & \rotatebox{90}{Pickup} & \rotatebox{90}{Camping} & \rotatebox{90}{Truck} & \rotatebox{90}{Other} & \rotatebox{90}{Tractor} &  \rotatebox{90}{Boat} & \rotatebox{90}{Van}  & $\text{mAP}_{50}$(\%)$\uparrow$  \\
        \midrule
        ZAQ\textcolor{red}{[CVPR'21]}~\cite{9577919} & 88.0 & 85.8 & 70.5 &  \underline{79.4} & 45.1 &  \underline{88.1} & 67.7 & \underline{84.0} & 76.1  \\
        AFD\textcolor{red}{[AAAI'21]}~\cite{ji2021show} & 88.5 & 85.5 & 71.3 & 73.5  & 58.7 & \cellcolor{tableyellow}\textbf{ 89.1}  & 59.7 & 80.4 &  75.8 \\
        ReviewKD\textcolor{red}{[CVPR'21]}~\cite{chen2021distilling} & 85.1 & 84.5 & 72.8 & 73.6  & 58.4 & 84.3  & 68.6 & \cellcolor{tableyellow}\textbf{94.0} &  77.7 \\
        YOLOFusion\textcolor{red}{[PR'22]}~\cite{qingyun2022cross} & \underline{91.7} & \underline{85.9} & 78.9 & 78.1  & 54.7 & 71.9  & 71.1 & 75.2 &  \underline{78.6} \\
        SuperYOLO\textcolor{red}{[TGRS'23]}~\cite{10075555} & 91.1 & 85.7 & \underline{79.3} & 70.2 & 57.3 &  80.4 & 60.2 & 76.5 & 75.9 \\
        OST\textcolor{red}{[TGRS'23]}~\cite{zhang2023guided} & 91.1 & \cellcolor{tableyellow}\textbf{87.7} & 74.9 &\cellcolor{tableyellow}\textbf{82.2}& \underline{64.6} & 84.9 & 60.2 & 82.9& \underline{78.6} \\
        DMM\textcolor{red}{[arxiv'24]}~\cite{zhou2024dmm} & 84.2 & 78.8 & 79.0 & 65.7 & 56.2 &  72.3 & \underline{72.3} & 72.5 &  75.0 \\
        ICAFusion\textcolor{red}{[PR'24]}~\cite{shen2024icafusion} & - & - & - &  - & - &  - & - & - & 76.6  \\
        % TMSI-Net\textcolor{red}{[TGRS'24]}~\cite{10364853} & 93.9 & 69.1 & 75.4 &  46.4 & 43.1 &  84.0 & 43.4 & 51.7 & 67.4  \\
        C\textsuperscript{2}Former-S\textsuperscript{2}ANet\textcolor{red}{[TGRS'24]}~\cite{yuan2024c}  & 76.7 & 68.7 & 63.2 & 52.0 & 41.9 &  59.8 & 43.3 & 48.0 & 55.6  \\  
        Ours & \cellcolor{tableyellow}\textbf{91.8} & 69.3 & \cellcolor{tableyellow}\textbf{82.3} & 78.5 & \cellcolor{tableyellow}\textbf{84.6} &85.8 & \cellcolor{tableyellow}\textbf{75.6} & 81.6 & \cellcolor{tableyellow}\textbf{81.2}\\
        \bottomrule
    \end{tabular} 
     }
     \caption{Mean Average Precision (mAP) comparison of different methods on the VEDAI dataset. The best result is shown in \textbf{bold} and the second best is shown with \underline{under line}.}
    \label{tbl:table2}
      \vspace{-10pt}
\end{table}

As shown in~\cref{tbl:table1}, multispectral (IR+RGB) fusion on the DroneVehicle dataset improves detection accuracy compared to single-modality (IR or RGB). Our method, as highlighted in~\cref{tab:sizespeed}, excels in accurately identifying small targets, such as cars, within the mixed-size target DroneVehicle dataset. It outperforms MGMF~\cite{10659747} with a 6.07 $\times$ speedup and a 27.29 $\times$ reduction in model size. While it exhibits a marginally lower overall accuracy, it achieves superior performance specifically for small cars. This indicates it also demonstrates powerful adaptability in scenarios with mixed-size targets, while ensuring excellent real-time performance and a lightweight model. 
% In addition, by comparing IR modality, RGB modality, and IR+RGB modality on the DroneVehicle dataset, we observe that multi-spectral fusion effectively improves detection accuracy, as shown in~\cref{tbl:table1}. Our method demonstrates superior performance on small targets like cars. Additionally, it shows competitive performance in mixed-size target scenarios, indicating its adaptability to diverse detection tasks. 

% As~\cref{tab:sizespeed} shows, our method accurately recognizes small targets on the mixed size target DroneVehicle dataset, while maintaining good real time performance and model lightness.  It is 6.07 $\times$ faster and 27.29 $\times$ smaller than MGMF~\cite{10659747}. While its recognition accuracy is slightly lower than MGMF's overall, it performs best for small target cars. 

\definecolor{ncar}{RGB}{255, 0, 0}
\definecolor{ntrunk}{RGB}{255, 192, 203}
\definecolor{nbus}{RGB}{255, 120, 30}
\definecolor{nvan}{RGB}{255, 200, 120}
\definecolor{nFreightCar}{RGB}{200, 220, 0}
\begin{table}[ht]
    \centering
    \renewcommand\arraystretch{1.1}
    \resizebox{\columnwidth}{!}{ % Fit to single-column width
     %\scalebox{0.9}{
    \begin{tabular}{c|c|ccccc|c}
        \toprule
         Method & Modality& \rotatebox{90}{\textcolor{ncar}{$\blacksquare$} Car} & \rotatebox{90}{\textcolor{ntrunk}{$\blacksquare$}Truck} & \rotatebox{90}{\textcolor{nbus}{$\blacksquare$}Bus} & \rotatebox{90}{\textcolor{nvan}{$\blacksquare$}Van} & \rotatebox{90}{\textcolor{nFreightCar}{$\blacksquare$}FreightCar} & $\text{mAP}_{50}$(\%)$\uparrow$ \\
        \midrule
        Oriented RepPoints~\cite{li2022oriented} & \multirow{4}{*}{IR} & 89.9 & 55.6 & 89.1 & 48.1 & \underline{57.6} & 68.0 \\
       S\textsuperscript{2}ANet~\cite{han2021align} & & 89.9 & 54.5 & 88.9 & 48.4 & 55.8 & 67.5\\
        LSKNet-OBB~\cite{li2023large} & & \underline{90.3} & \cellcolor{tableyellow}\textbf{73.3} & 89.2 & \underline{53.2} & \cellcolor{tableyellow}\textbf{57.8} & \cellcolor{tableyellow}\textbf{72.8} \\
        ReDet~\cite{han2021redet} & & 90.0 & 61.5 & \underline{89.5} & 46.6 & 55.6 & 68.6\\
         Ours & & \cellcolor{tableyellow}\textbf{96.8} & \underline{68.3} & \cellcolor{tableyellow}\textbf{93.9} & \cellcolor{tableyellow}\textbf{59.5} & 44.0 & \underline{72.5}\\
        % Mambayolo~\cite{wang2024mamba} & & \cellcolor{tableyellow}\textbf{97.4} & \underline{71.0} & \cellcolor{tableyellow}\textbf{94.9} & \cellcolor{tableyellow}\textbf{63.1} & 57.5 & \cellcolor{tableyellow}\textbf{76.8} \\
        \midrule
        Oriented RepPoints~\cite{li2022oriented} & \multirow{4}{*}{RGB} & 84.4 & 55.0 & 85.8 & 46.6 & 39.5 & \underline{62.3}\\
        S\textsuperscript{2}ANet~\cite{han2021align} & & 80.0 & 54.2 & 84.9 & 43.8 & \underline{42.2} & 61.0\\
        LSKNet-OBB~\cite{li2023large} & & \underline{89.5} & \underline{70.0} &  \underline{89.4} & 56.9 & \cellcolor{tableyellow}\textbf{51.8} & \cellcolor{tableyellow}\textbf{71.5}\\
        ReDet~\cite{han2021redet} & & 69.5 & 47.9 & 31.0 & \cellcolor{tableyellow}\textbf{77.0} & 29.0 & 51.0\\
        Ours & & \cellcolor{tableyellow}\textbf{94.6} & \cellcolor{tableyellow}\textbf{70.5} & \cellcolor{tableyellow}\textbf{94.2} & \underline{61.3} & 36.9 & \cellcolor{tableyellow}\textbf{71.5}\\
        % Mambayolo~\cite{wang2024mamba} & & \cellcolor{tableyellow}\textbf{94.6} & \cellcolor{tableyellow}\textbf{70.5} & \cellcolor{tableyellow}\textbf{94.2} & \underline{61.3} & 36.9 & \cellcolor{tableyellow}\textbf{71.5} \\
        \midrule
        DPDETR~\cite{guo2024dpdetr} & \multirow{9}{*}{\makecell{RGB \\ +IR}} & 90.3 & \underline{78.2} & 90.1 & 64.9 & \underline{75.7} & \underline{79.8}\\
        DMM~\cite{zhou2024dmm} & & 90.4 & \cellcolor{tableyellow}\textbf{79.8} & 68.2 & \cellcolor{tableyellow}\textbf{89.9} & 68.6 & 79.4 \\
        % CIAN\cite{zhang2019cross} & & 90.1 & 63.8 & 89.1 & 50.3 & 60.7  & 70.8 \\
       S\textsuperscript{2}ANet~\cite{han2021align} & & 90.0 &64.5  & 88.2 & 53.2 & 61.7 &  71.5\\
       % MBNet\cite{zhou2020improving} & & 90.1 & 64.4 & 88.8 & 53.6 & 62.4 & 71.9  \\
       % AR-CNN\cite{zhang2021weakly} & & 90.1 &  64.8& 89.4 & 51.5 & 62.1 & 71.6  \\
        TSFADet~\cite{yuan2022translation} & & 89.9 & 67.9 & 89.9 & 54.0 & 63.7 & 73.1 \\
        %E2E-MFD\cite{zhang2024efficientmfd} & & 90.3 & 79.3 & 89.8 & 63.1 & 64.6 & 77.4  \\
        C\textsuperscript{2}Former-S\textsuperscript{2}ANet~\cite{yuan2024c} & & 90.2 & 68.3 & 89.8 & 58.5 & 64.4 & 74.2 \\
         UA-CMDet~\cite{9759286} & & 87.5& 60.7 & 87.1 & 37.9 & 46.8  & 64.0  \\
         MKD~\cite{huang2023multimodal} & & \underline{93.5}&62.5  & \underline{91.9} & 44.5 &  52.7 & 69.0 \\
         MGMF~\cite{10659747} & & 91.4& 70.1 & 91.1 & \underline{69.4} & \cellcolor{tableyellow}\textbf{78.5}  &\cellcolor{tableyellow}\textbf{80.2}  \\
        Ours & & \cellcolor{tableyellow}\textcolor{red}{\textbf{96.8}} & 69.2 & \cellcolor{tableyellow}\textbf{93.1} & 60.4 & 52.8 & 74.5 \\
        \bottomrule
    \end{tabular} 
     }   
       \caption{mAP comparison of different methods on the DroneVehicle dataset. The target pixel sizes are sorted as: \textcolor{red}{\textbf{Car}} \textless Bus \textless Van \textless Freight Car \textless Truck. For the smallest targets like cars, ours has better performance than the CNN architecture (UA-CMDet~\cite{9759286}, MKD~\cite{huang2023multimodal}), Transformer architecture (DPDETR~\cite{guo2024dpdetr}), and other models based on Mamba architecture (DMM~\cite{zhou2024dmm}, MGMF~\cite{10659747}).}
    \label{tbl:table1}
    \vspace{-10pt}
\end{table}

% \cref{tab:sizespeed}表明，在DroneVehicle大小目标混合的数据集上，我们的方法展现出了显著优势与特点。与MGMF~\cite{10659747}相比，我们的方法在速度上提升了6.07倍，模型大小减少了27.29倍。在识别精度上，虽略低于MGMF，但聚焦于该数据集中的car小目标类别时，我们的方法表现最优。 这表明我们的方法在处理小目标时实现高精度识别的同时保持了较好的实时性与模型轻量化。
% As shown in~\cref{tab:sizespeed}, our method demonstrates remarkable advantages and characteristics on the DroneVehicle dataset with a mixture of large and small targets. Compared to MGMF~\cite{10659747}, our speed is 6.07$\times$faster and model size is 27.29$\times$smaller. In terms of recognition accuracy, although it is slightly lower than that of MGMF, our method performs the best when focusing on the small target category of cars in this dataset. This indicates that our method can achieve high-precision recognition for small targets while maintaining good real-time performance and model lightweighting. 

\begin{table}
\setlength{\tabcolsep}{0.01\linewidth}
\centering
\renewcommand{\arraystretch}{1.1}
\small
\scalebox{0.8}{
\begin{tabular}{c|c|c}
 \toprule
 Method & Size (MB)$\downarrow$ & Speed (Fps)$\uparrow$   \\
 \midrule
 DPDETR~\cite{guo2024dpdetr} & 90.10& -  \\
                             DMM~\cite{zhou2024dmm} & \underline{87.97}& -  \\
                             S\textsuperscript{2}ANet~\cite{han2021align} & 142.80& - \\
                             TSFADet~\cite{yuan2022translation} & 104.70 & 18.60  \\
                             C\textsuperscript{2}Former-S\textsuperscript{2}ANet~\cite{yuan2024c}  & 120.80& -  \\
                             UA-CMDet~\cite{9759286} & 234.00 & 9.12 \\
                             MKD~\cite{huang2023multimodal} & 242.00& \underline{42.30}  \\
                             MGMF~\cite{10659747} & 465.92& 6.10 \\
                             Ours &  \cellcolor{tableyellow}\textbf{ 17.07}&  \cellcolor{tableyellow}\textbf{43.10} \\                           
\bottomrule
\end{tabular}
}
\caption{Comparison of Frames Per Second (FPS) and parameter sizes of different methods on the DroneVehicle dataset. The methods involved correspond to \cref{tbl:table1}.}
\label{tab:sizespeed}
 \vspace{-10pt}
\end{table}
%%%%%%%%%%%%%%%%%%%%%%%%%%%%%%%%%%%%%%%%%%%%%%%%%%%%%%%%%%%%%%%%%%%%%%%%%%%%%%%%%%%%%%%%%%%%%%%

%在不同场景下，可视化实验对比图以及红外和可见光模态下的groundtruth。我们选取了具有代表性的5种场景:1.光线分布不均的夜晚场景（同时存在光线过曝和光线过暗）——a;2.光线充足但有树木遮挡背景复杂的白天——b/e;3.由于高速采集导致目标模糊的夜晚场景，且同时包含大尺寸目标和小目标;——c4.多目标集中且有遮挡的夜晚场景——f;5.背景同时包含与目标相似物体的夜晚场景——g/d。

% 以CNN为主要结构的YOLOv8在（a）-（g）7张图片上均有检测错误。以Transformer为主要结构的XX在（b）和（d）-（f）4张图片上有检测错误。ours以mamba为主要结构仅在2张图片上有检测错误。经分析，（d）是由于该图中的freight car与car外形相似，并且多个目标相对集中。（f）是由于有树木遮挡，且为高速采集状态下的图片，图像本身就比较模糊。

\begin{figure}[!ht]
  \centering
  % \vspace{-8pt}
  \includegraphics[width=1\linewidth]{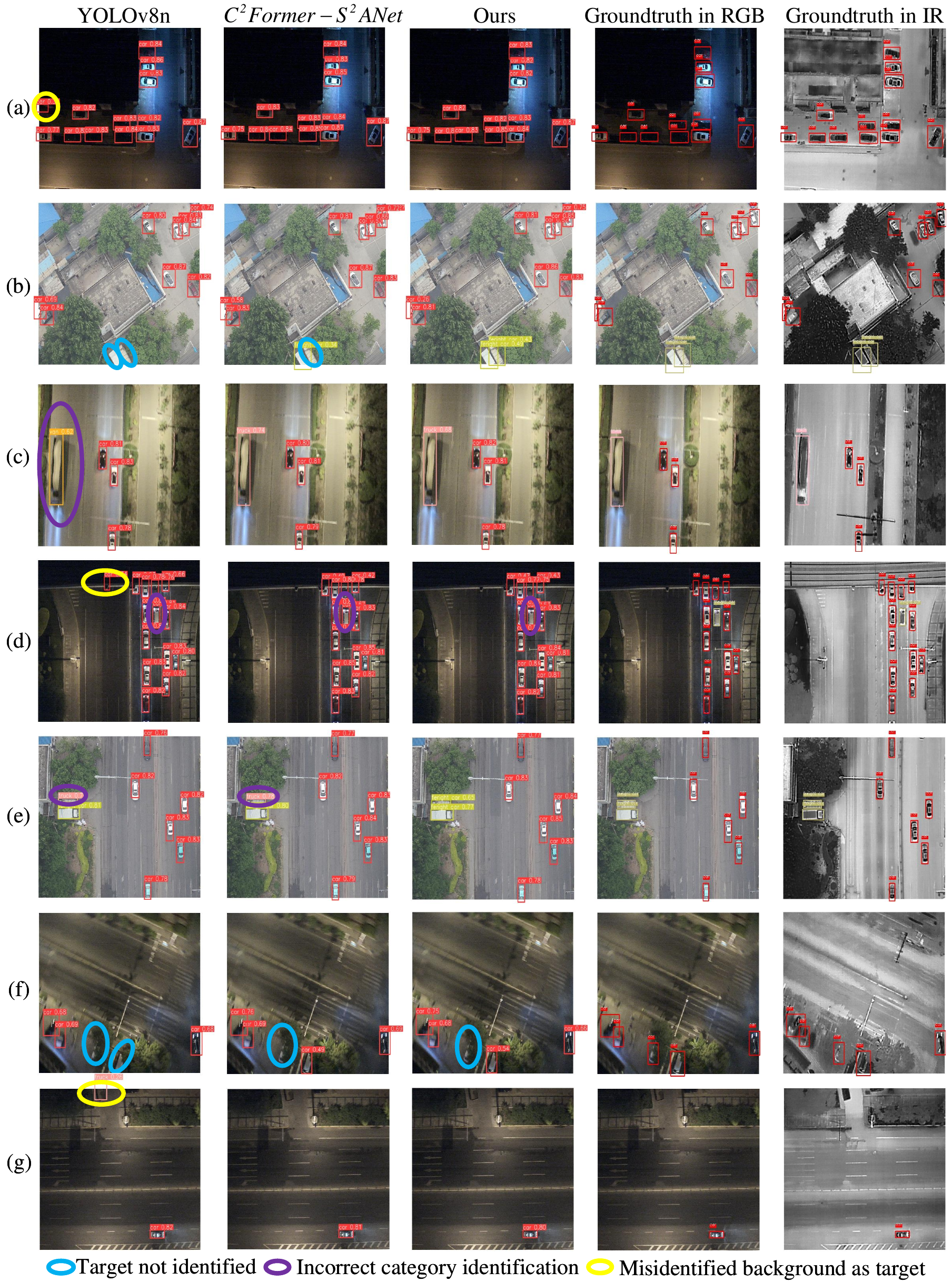}
%可视化对比实验
% \includegraphics[width=1\linewidth]{sec/fig/可视化对比实验.png}
   \caption{Visualization of experimental comparison plots and ground truth in infrared and visible modes in different scenarios. Specifically, we have selected five representative scenarios. \textbf{(a):} Night scenes with uneven light distribution (both overexposed and low light); \textbf{(b) and (e):} Daytime with plenty of light but complex backgrounds shaded by trees; \textbf{(c):} Night scenes with blurred targets due to high-speed acquisition and containing both large and small targets; \textbf{(f):} Multi-target and occluded night scenes; \textbf{(g) and (d):} Night scenes where the backgrounds also contain objects similar to the targets.}
   \label{fig:1}
    % \vspace{-1em}
\end{figure}

%[进一步验证可靠性] [强调本文设计的模型在这些模型中的优势] [横向和纵向比较］
As shown in~\cref{fig:1}, YOLOv8n with CNN as the main structure has detection errors on all \textbf{7} images (a)-(g). C\textsuperscript{2}Former-S\textsuperscript{2}ANet~\cite{yuan2024c} with Transformer as the main structure has detection errors in (b) and (d)-(f) on the \textbf{4} images. Ours only has detection errors on \textbf{2} images. On analysis, (d) is due to the similarity in shape of the Freight Car and Car in this image and the relative concentration of multiple targets. (f) is a picture in a high-speed acquisition state where the image itself is blurred and obscured by trees. %{\color{red}[Further verification of reliability] [Emphasize the advantages of the model designed in this paper among these models] }

%【给出检测结果图，不同算法相同的图，本文方法的优缺点，缺点！】也要进行总结。图7各种算法的结果以可视化的方式呈现，其中真阳性（TP）实例用绿色方框表示，假阳性（FP）和假阴性（FN）实例分别用蓝色和红色方框表示。这种可视化技术有助于清楚地理解算法结果。[对图中数据的解释]。
%XXX.{\color{red}[Give the detection results graph, different algorithms the same graph, the advantages and disadvantages of this paper's method, disadvantages! are also to be summarized. INSERT FIGURE 7 The results of various algorithms are presented by visualization, where true positive (TP) instances are represented by green boxes and false positive (FP) and false negative (FN) instances are represented by blue and red boxes, respectively. This visualization technique helps to clearly understand the algorithm results.] [Interpretation of data in the figure].}

%end<待办3：>填写下面的空，并且把红色标题融进去

%begin<待办5：>填写下面的空，并且把红色标题融进去

%【模仿上面的】

\subsection{Ablation Study}

% \begin{figure}[t]
%   \centering
%   %\fbox{\rule{0pt}{2in} \rule{0.9\linewidth}{0pt}}
%    \includegraphics[width=0.8\linewidth]{sec/fig/消融实验.png}

%    \caption{xxx}
%    \label{fig:1}
% \end{figure}

% \begin{figure}[t]
%   \centering
%   %\fbox{\rule{0pt}{2in} \rule{0.9\linewidth}{0pt}}
%    \includegraphics[width=0.8\linewidth]{sec/fig/消融实验现有结果.png}

%    \caption{xxx}
%    \label{fig:1}
% \end{figure}
%我们首先分析 xxx模块的效果，因为它提供了xxx，这决定了模型的xxx。如表XXX第X行和第X行所示。加入xxx模块后，基线模型的性能提高了x%。它验证了XX和通过XX增加XX灵活性对于检测任务都是有效的。此外，当集成这两个子模块时，模型的性能相对于基线提高了 XX%，这表明可以通过减少XXX来提高XXX的质量。

\begin{figure}
    \begin{center}
        \includegraphics[width=0.48\textwidth]{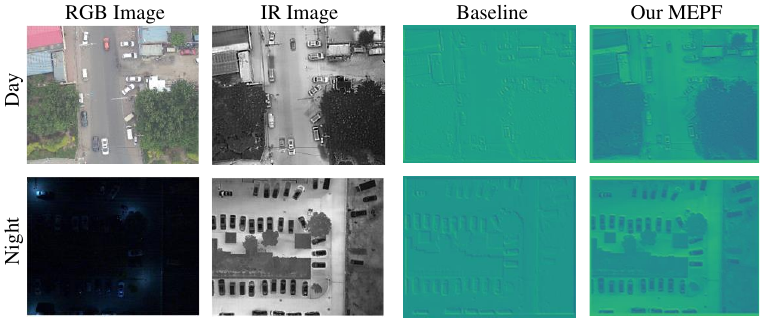}
        \caption{Comparison of feature map visualization between our MEPF module and the baseline pixel-level fusion method under different lighting conditions.
        } 
        % 不同光照条件下我们的MEPF像素级融合方法和baseline像素级融合方法的特征图可视化比较。
        \label{fig:MEPFMODULE}
    \end{center}
    \vspace{-1em}
\end{figure}

%Effect of XXX module. 
%我们首先分析了MEPF模块的效果，因为它提供了更精细的像素级特征融合方法，这直接决定模型的精确度。如XXX第一行和第二行所示。使用MEPF模块后，基线模型的性能提高了+2.9%，这表明我们的MEPF可以提高可见光模态和红外模态融合的效果。其中，基线模型是传统的像素级融合方法与yolov8n的结合。
% \subsubsection*{}
\noindent{\textbf{Effect of MEPF module.}}
% % We first analyze the effect of the MEPF module since it offers a finer pixel-level feature fusion method, directly determining the model's accuracy. As shown in the first and second lines of~\cref{tbl:table3}. The performance of the baseline model is improved by +2.9\% with the MEPF module, which shows that our MEPF can improve the fusion of visible and infrared modes. In this case, the baseline model is a combination of traditional pixel-level fusion methods~\cite{10075555} and yolov8n. As shown in~\cref{fig:MEPFMODULE}.
% We first analyze the effect of the MEPF module, which provides a finer pixel-level feature fusion method. As shown in~\cref{tbl:table3} (first and second lines), the MEPF module improves the baseline model's performance by +2.9\%, demonstrating its effectiveness in fusing visible and infrared modalities. The baseline model combines traditional pixel-level fusion methods~\cite{10075555} with YOLOv8n, as illustrated in~\cref{fig:MEPFMODULE}. Our MEPF module generates higher-quality fusion images by effectively integrating visible and infrared spectral information, particularly in challenging lighting scenarios such as low light or overexposure.
Our MEPF module can produce a higher quality fusion image. As shown in~\cref{tbl:table3} (first and second lines), it improves the baseline model's performance by +2.9\%. The baseline model consists of three components: the VMamba~\cite{liu2024vmamba} backbone, traditional pixel-level fusion~\cite{10075555}, and the YOLOv8n head~\cite{yolov8_ultralytics}, as illustrated in~\cref{fig:MEPFMODULE}. Our MEPF module generates higher-quality fusion images, particularly in challenging lighting conditions like low light or overexposure.
%然后我们验证了XX模块解决XXX问题的必要性。结果表明，尽管 XXX在没有XXXX指导的情况下在模型训练中稳定性较差，但将XXX纳入基线可以提供 XXX% 的显着改进。这是由于XX调整XXX以获得RGB图像中的可靠信息。
% \subsubsection*{Effect of ESTD module}
%然后我们验证了ESTD模块解决小目标漏检和误检问题的必要性。结果表明，将ESTD纳入基线，可以提供+3.3%的显著改进。这是由于局部注意力可以更好的捕捉小目标信息。

\begin{figure}
% \vspace{-8pt}
    \begin{center}
        \includegraphics[width=0.48\textwidth]{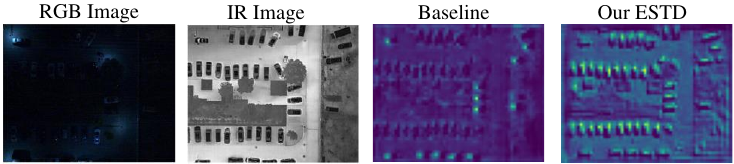}
        \caption{Comparison of feature map visualizations for our ESTD module and baseline approach. 
        } 
        % 我们的ESTD module和baseline方法的特征图可视化比较。
        \label{fig:MEPFMODULE22}
    \end{center}
    \vspace{-1em}
\end{figure}

\noindent{\textbf{Effect of ESTD module.}}
% Then, we verify the effectiveness of the ESTD module in solving the problem of small target omission and error detection. The results show that including ESTD in the baseline provides a significant improvement of +3.3\%. This is due to the increased ability to capture local details. As shown in~\cref{fig:MEPFMODULE22}.
% Then, we verify the effectiveness of the ESTD module. By augmenting Mamba's global attention with local attention mechanisms, the ESTD module significantly improves the model's ability to capture fine-grained details, which are often missed by global attention alone. This is particularly critical for small targets, as their limited size and weak features make them prone to omission and false detection in complex backgrounds. As shown in~\cref{fig:MEPFMODULE22}, incorporating the ESTD module into the baseline improves performance by +3.3\%, demonstrating its effectiveness in enhancing small target detection accuracy.
The ESTD module improves fine-grained detail capture through its specialized block, which adapts the Mamba architecture to optimize receptive fields for small target detection by strengthening local attention mechanisms. As shown in~\cref{fig:MEPFMODULE22}, incorporating the ESTD module into the baseline improves performance by +3.3\%, demonstrating its effectiveness in enhancing small target detection accuracy.
%这是因为增强了局部细节的捕捉能力。
% \subsubsection*{Effect of CARG module}
%我们验证了CARG模块同样重要。结果表明，将CARG纳入基线，可以提供+3.4%的显著改进。这是由于通道注意力机制可以更好聚焦图像中有用的信息，而忽略复杂背景干扰。

\begin{figure}
% \vspace{-8pt}
    \begin{center}
        \includegraphics[width=0.48\textwidth]{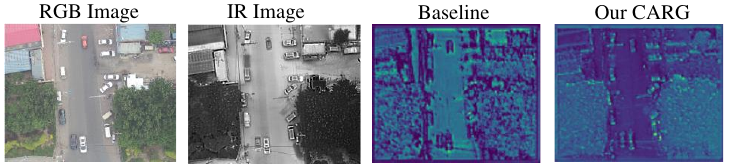}
        \caption{Comparison of feature map visualizations for our CARG module and the baseline approach. 
        } 
        % 我们的CARG module和baseline方法的特征图可视化比较。
        \label{fig:MEPFMODULE33}
    \end{center}
    \vspace{-1em}
\end{figure}

\noindent{\textbf{Effect of CARG module.}}
% We have verified that the CARG module is equally important. The results show that including CARG in the baseline provides a significant improvement of +3.4\%. This is because the channel attention mechanism can focus on useful information in the image better while ignoring complex background interference. As shown in~\cref{fig:MEPFMODULE33}.
Our CARG module can better distinguish between complex backgrounds and targets. By assigning higher weights to key features of small targets and suppressing background channels, the CARG module boosts the model's anti-interference ability. As shown in~\cref{fig:MEPFMODULE33}, incorporating the CARG module into the baseline improves performance by +3.4\%.
%我们还验证了同时应用MEPF模块ESTD模块以及CARG模块的效果比以往都要好。相比于基线可以提供+3.6%的显著改进。
% We also verify that the simultaneous application of the MEPF module, ESTD module, and CARG module is more effective than ever. Compared to the baseline, our method provides a significant improvement of +3.6\%.

% In conclusion, not only each module but also the whole proposed framework can contribute to multimodal target detection under small target conditions.

\begin{table}
\small
    \centering
    \renewcommand{\arraystretch}{1.1}
    %\setlength{\tabcolsep}{3pt} % Adjust column spacing to reduce table width
   % \resizebox{\columnwidth}{!}{ % Fit to single-column width
   \scalebox{0.85}{
    \begin{tabular}{l|ccc|c|c}
        \toprule
         Method & MEPF & ESTD & CARG  & $\text{mAP}_{50}$(\%)$\uparrow$   & Parameter (MB) $\downarrow$ \\
        \midrule
       
        \uppercase\expandafter{\romannumeral1} & & & &   70.9 & \textbf{11.49} \\ 
      \uppercase\expandafter{\romannumeral2} & \checkmark & & & 73.8 & 11.50 \\
      \uppercase\expandafter{\romannumeral3} & \checkmark & \checkmark & & 74.2 & 16.61 \\
     \uppercase\expandafter{\romannumeral4} & \checkmark & & \checkmark &  74.3 & 16.92 \\
        \uppercase\expandafter{\romannumeral5} &\checkmark & \checkmark & \checkmark & \textbf{74.5} & 17.07\\
       
        \bottomrule
    \end{tabular} 
     }
     \caption{Ablation study on DroneVehicle. The best result is shown in \textbf{bold}. \uppercase\expandafter{\romannumeral1}: Baseline; \uppercase\expandafter{\romannumeral2}: Baseline+MEPF; \uppercase\expandafter{\romannumeral3}: Baseline+MEPF+ESTD; \uppercase\expandafter{\romannumeral4}: Baseline+MEPF+CARG; \uppercase\expandafter{\romannumeral5}: Baseline+MEPF+ESTD+CARG.}
    \label{tbl:table3}
     \vspace{-10pt}
\end{table}

%\cref{tbl:SA25}表明，尽管大内核在车辆单一目标识别上有微弱优势，但综合考虑整体检测精度与参数规模，1×1 内核在平衡模型性能与复杂度上表现最佳。这也正是我们最终采用 1×1 内核的原因.
As shown in~\cref{tbl:SA25}, larger kernels have a slight superiority in single vehicle recognition. However, considering overall accuracy and parameter size, the 1$\times$1 kernel best balances performance and complexity. Thus we choose it.

\begin{table} 
\small  
\centering 
% \caption{Effect of Different Activation Functions on Recognition Accuracy (mAP ↑) of IASPP Modules}  
% \label{tab:2}  
\scalebox{1}{ 
\begin{tabular}{llll}  
\toprule  
Kernel size      & Car & $\text{mAP}_{50}$(\%)$\uparrow$ & Parameter $\downarrow$\\   
\hline  
1 $\times$ 1           &96.8&\textbf{74.5}&\textbf{4448887} \\ 
3 $\times$ 3           &97.0&73.2&4449008 \\ 
7 $\times$ 7           &\textbf{97.4}&73.8&4449488 \\ 

\bottomrule  
\end{tabular}}  
\caption{Ablation on Spatial Attention Kernel Size.}
\label{tbl:SA25}
\vspace{-10pt}
\end{table}
%XXX.{\color{red}[Focus: Evaluating the impact of each improvement module on the overall network][Emphasis under the same configuration]}

%XXX.{\color{red}[Corresponding to Abstract 1 - Version 2.0] [Corresponding to Abstract 2 - Version 3.0] [Corresponding to summary 3 - version 4.0][Comparison of the original final, i.e. 1.0 and 4.0]}

%XXX.{\color{green}[Insert Table 6 Ablation experiments based on the dronevehicle dataset.] [Insert Table 6 Ablation experiments based on the dronevehicle dataset.] [Parameters GFLOPs mAP (\%)]}
%end<待办5：>填写下面的空，并且把红色标题融进去

%begin<待办6：>如果可能的话分析一下我们的方法有哪些限制【类似下面的方式】

% %6.4. 限制
% 我们的方法采用高斯滤波器作为箱式滤波器的近似值来提高效率。但是，这种近似会引入误差，尤其是当 Gaussian 的屏幕空间较小时。这个问题与我们的实验结果相关，其中增加的缩小会导致更大的错误，如表 2 所示。此外，训练开销略有增加，因为每个 3D 高斯的采样率必须每 m = 100 次迭代计算一次。目前，这种计算是使用 PyTorch [35] 执行的，更高效的 CUDA 实现可能会减少这种开销。设计一个更好的数据结构来预计算和存储采样率，因为它完全取决于相机的姿势和本质，是未来工作的一个途径。如前所述，采样率计算是训练过程中唯一的先决条件，3D平滑滤波器可以根据Eq. 9与高斯原语融合，从而消除渲染过程中的任何额外开销。

%我们的方法采用MEPF和ESTVSS Block增强小目标的识别准确度，并且仅有17.07MB参数量。但是，这种轻量化的方法在较大目标的检测上效果一般。如\cref{tbl:table1}中所示，我们在DroneVehicle数据集上总体精度没有DMM方法高，而在VEDAI数据集上臂DMM方法高。仔细分析可知，在DroneVehicle数据集上DMM方法主要在Trunk和Van较大的目标上有更好的效果。而我们的方法主要在car这类小目标上有较高的精度。这也说明了为什么我们会在VEDAI数据上有更好的效果，因为VEDAI数据集的目标均值更小。设计一个同时可以精确检测大尺寸目标和小目标的模型是未来工作的一个途径。
\subsection{Limitations}
Our method, combining the MEPF and ESTVSS Block, achieves high small target detection accuracy with only 17.07MB parameters. However, it is less effective for larger targets, as shown in~\cref{tbl:table1}, where our overall accuracy on the DroneVehicle dataset lags behind DMM~\cite{zhou2024dmm}. This is because DMM excels at detecting larger targets like Trucks and Vans, while our method is optimized for small targets like cars. On the VEDAI dataset, which has a smaller target size (as shown in~\cref{fig:size}), our method outperforms DMM. Future work will focus on designing a model capable of accurately detecting both large and small targets.

\section{Conclusion}

%begin<待办1：>进行修改——把下图的内容填空到下面填空中，以及几个红色的标题也要融入到段落中

% 我们【设计】了【MM-ViMY】，这是一种通过 【引入（可见光+红外）像素级融合（创新点1）+主干网络引入VssBlock（创新点2）+head引入XssBlock（创新点3）】改进 【Vision mamba】的技术，用于【小目标检测】。我们的 【创新点1】有效地【融合】了【可见光的纹理和色彩信息以及红外的XXX信息】，以【实现模态互补】，而 【创新点2】则近似于【mamba文本token输入XXX】以模拟【XXX】过程。当【对小目标】进行测试时，【MM-ViMY】在【无人机远距离复杂背景检测XX】场景中的性能明显优于最先进的方法，从而更好地【xxx】。

%%我们提出了$S_{6}^{4}$-MSTD模型，这是一个将像素级可见光与红外融合模块与改进后的mamba视觉检测主干网络相结合的小目标检测模型。我们的掩码增强像素级融合模块MEPF有效的融合了可见光图像的纹理和色彩信息以及红外图像的热辐射信息，从而补充了小目标的特征。此外，我们在Vmamba的基础上引入增强小目标检测模块ESTD和卷积注意残差门模块CARG来优化小目标特征提取和表达能力。最后我们在DroneVehicle和VEDAI数据集上进行了评估，验证了我们的方法具有参数规模更小以及准确度高的性能优势。

% We propose the $S_{6}^{4}$-MSTD model, aiming for a small target detection model that combines a pixel-level visible and infrared fusion module with a modified Mamba visual detection backbone network. Our MEPF module effectively fuses texture and color information from visible images and thermal radiation information from infrared images to complement small target features. In addition, we introduce the ESTD module and the CARG module to optimize the small target feature extraction and representation capability. Finally, to demonstrate the efficacy of our approach, we conduct extensive evaluations on the DroneVehicle and VEDAI datasets, validating the performance advantages of our method with smaller parameter sizes as well as high accuracy.%{\color{red}[To be done: populate the results of the experiment][Compared to XX, our method is faster, has smaller parameter sizes, lower arithmetic requirements, and little loss of accuracy.]}
%end<待办1：>进行修改——把下图的内容填空到下面填空中，以及几个红色的标题也要融入到段落中

We propose the $S_{6}^{4}$-MSTD model, a novel selective structured state space model designed for multispectral-fused small target detection. Our approach introduces the MEPF module for efficient multispectral fusion, which achieves superior fusion quality with merely 1650 parameters. To further enhance the Mamba-based model's capability in small target detection, we introduce the ESTD and CARG modules, significantly improving the backbone's ability to capture subtle clues of small targets. Extensive experiments on the DroneVehicle and VEDAI datasets demonstrate the effectiveness of our method, which maintains an optimal balance between detection accuracy and suitability for lightweight deployment on edge devices.

%%%%%%%%%%%%%%%%%%%%%%%%%%%%%%%%%%%%%%%%%%%%%%%%%%%%qqz old%%%%%%%%%%%%%%%%%%%%%%%%%%%%%%%%%%%%%%%%%%%%%%%%%%%%%%%%

% \section{Conclusion}

% XXX.{\color{red}[Summarize the algorithm in this paper: a xxx algorithm for xxx scenarios]}

% XXX.{\color{red}[Abstract 1]}

% XXX.{\color{red}[Abstract 2   -a different way of saying it, but it is the same as the abstract.]}

% XXX.{\color{red}[Abstract 3]}

% XXX.{\color{red}[Comparison of experimental performance in terms of: lightweighting metrics, performance]}

% XXX.{\color{red}[Comparison of experiments in: precision aspects of the index, the performance] }
%%%%%%%%%%%%%%%%%%%%%%%%%%%%%%%%%%%%%%%%%%%%%%%%%%%%qqz old%%%%%%%%%%%%%%%%%%%%%%%%%%%%%%%%%%%%%%%%%%%%%%%%%%%%%%%%

{
    \small
    \bibliographystyle{ieeenat_fullname}
    \bibliography{main}
}


\end{document}